\documentclass{article} 
\usepackage[final]{colm2026_conference}

\usepackage{microtype}
\usepackage{hyperref}
\usepackage{url}
\usepackage{booktabs}
\usepackage{amsmath}
\usepackage{amssymb}
\usepackage{graphicx}
\usepackage{multirow}
\usepackage[table]{xcolor}
\usepackage{longtable}
\usepackage{arydshln}
\usepackage{colortbl}
\usepackage{lineno}
\usepackage{adjustbox}
\usepackage{nicematrix}
\usepackage{subcaption}
\usepackage{wrapfig}
\usepackage{float}
\usepackage{subcaption}
\usepackage{algorithm}
\usepackage{algpseudocode}
\usepackage{hhline}
\usepackage{pifont}
\usepackage{fontawesome5}
\usepackage{bm}

\newcommand{\cmark}{\textcolor{green!60!black}{\ding{51}}}
\newcommand{\xmark}{\textcolor{red}{\ding{55}}}

\usepackage{lineno}
\newtheorem{proposition}{Proposition}
\newtheorem{theorem}{Theorem}
\definecolor{darkblue}{rgb}{0, 0, 0.5}
\hypersetup{colorlinks=true, citecolor=darkblue, linkcolor=darkblue, urlcolor=darkblue}

\title{\textsc{KronQ}: LLM Quantization via Kronecker-Factored Hessian}

\author{Donghyun Lee$^1$, Yuhang Li$^2$, Ruokai Yin$^2$, Priyadarshini Panda$^1$ \\
$^1$University of Southern California, $^2$Yale University\\
\texttt{donghyun.lee.1@usc.edu}\\[0.8em]
\centerline{\raisebox{-0.15em}{\large\faGithub}~~\url{https://github.com/Intelligent-Computing-Lab-Panda/KronQ}}\\
}

\newcommand{\KronQ}{\textsc{KronQ}}

\begin{document}

\ifcolmsubmission
\linenumbers
\fi

\maketitle

\begin{abstract}
\vspace{-3mm}
Post-training quantization (PTQ) is a widely adopted technique for compressing large language models (LLMs) without retraining. Most existing second-order PTQ methods, including GPTQ, construct quantization objectives from input activation statistics, effectively assuming that all output channels contribute equally to the layer-wise reconstruction objective. We propose \KronQ{}, a PTQ framework that challenges this assumption by introducing the gradient covariance into the quantization pipeline. Under the Kronecker-factored Hessian approximation, the quantization loss depends jointly on both the activation and gradient covariances, and \KronQ{} exploits this at two complementary levels. (1) \KronQ{} introduces bidirectional incoherence processing, extending the existing input-side random rotation to the output dimension using the gradient covariance, reducing weight magnitude variance across both input and output dimensions. (2) \KronQ{} derives a new sensitivity metric for inter-layer mixed-precision allocation, driven by the gradient and activation Hessian traces. Notably, in the case of 2-bit weight-only quantization on LLaMA-3-70B, while GPTQ and GPTAQ diverge or produce degenerate quantizations ($>$2000 perplexity on WikiText-2), \KronQ{} achieves 7.93 perplexity.
\vspace{-1mm}
\end{abstract}

\section{Introduction}
\vspace{-3mm}
Large language models (LLMs) have achieved remarkable performance across a wide range of natural language understanding and generation tasks~\citep{brown2020language,touvron2023llama,grattafiori2024llama}. However, their growing parameter counts, reaching hundreds of billions in recent models, pose significant deployment challenges, as the memory and computational requirements far exceed what is available on commodity and edge hardware. Quantization~\citep{gholami2022survey, liang2021pruning} provides an effective solution by compressing model weights and activations to low-bit representations, reducing both memory footprint and inference latency. Among quantization techniques, post-training quantization (PTQ) has attracted particular interest due to its practicality. It requires no retraining and operates on a pretrained model using only a small calibration dataset.

PTQ methods for LLMs span several  paradigms~\citep{zhao2025benchmarking}: (1) Compensation-based methods apply the Optimal Brain Surgeon (OBS) principle~\citep{hassibi1992second, lecun1989optimal} layer-by-layer, using the input activation covariance as a proxy Hessian to compensate for rounding error in weight quantization~\citep{frantar2022optimal, frantar2022gptq,kim2024boa,li2025gptaq, tseng2025model, kim2025guidedquant}. (2) Rotation-based methods apply orthogonal transforms to weight matrices and activations to suppress outliers before quantization~\citep{chee2023quip,tseng2024quip,ashkboos2024quarot,liu2024spinquant}. (3) Salience-based methods identify sensitive weight channels via activation magnitudes and protect them through per-channel scaling~\citep{lin2024awq,xiao2023smoothquant}. These paradigms are 
complementary and are often combined in practice.
In this work, we focus on the compensation-based family, in particular GPTQ~\citep{frantar2022gptq} and its successors, combined with orthogonal transforms from the rotation-based paradigm. These methods have become the dominant framework for LLM weight quantization and are widely integrated into production serving stacks, including HuggingFace Transformers~\citep{wolf-etal-2020-transformers}, GPTQModel~\citep{qubitium2024gptqmodel}, and vLLM~\citep{kwon2023efficient}.

\begin{figure*}
    \centering
    \includegraphics[width=\textwidth]{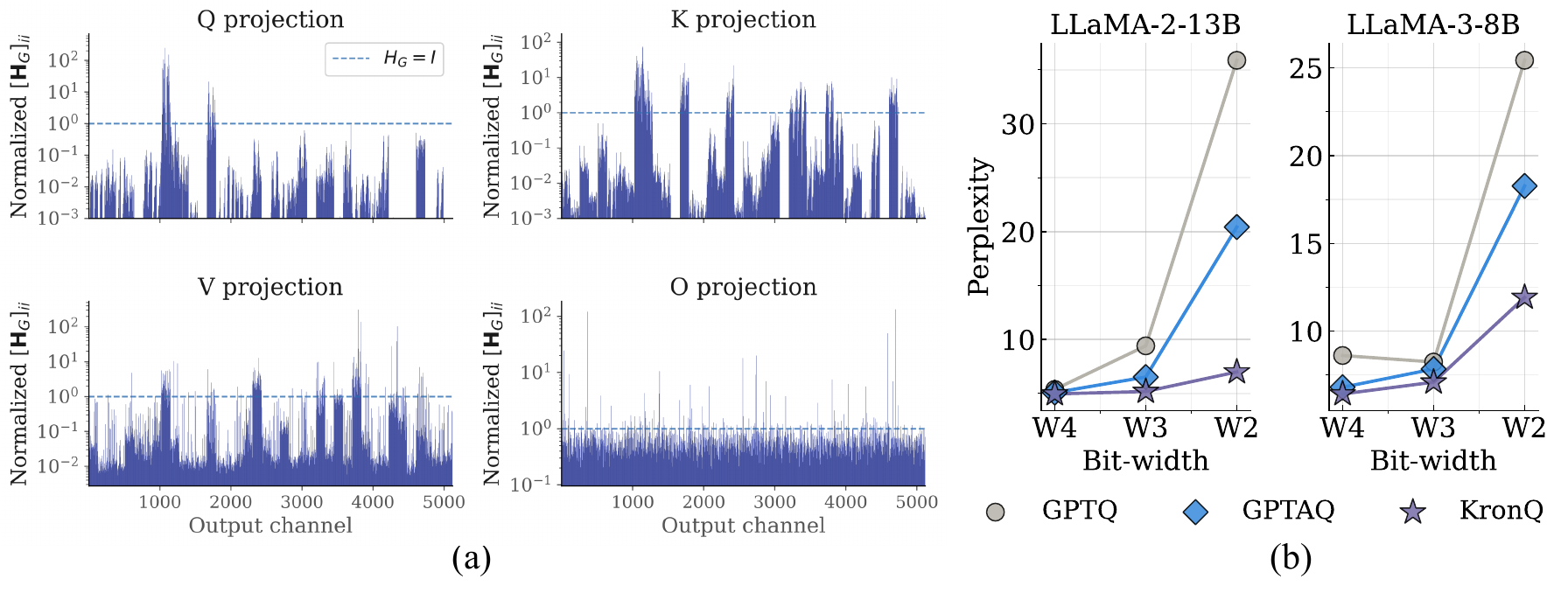}
    \vspace{-8mm}
    \caption{(a) Normalized diagonal entries of the gradient covariance $H_G$ for Q, K, V, and O projections in LLaMA-2-13B. Diagonal entries vary by orders of magnitude, revealing heterogeneous output-side sensitivity. (b) WikiText-2 perplexity of GPTQ, GPTAQ, and \KronQ{} on LLaMA-2-13B and LLaMA-3-8B across W4/W3/W2 weight-only quantization.}
    
    \label{fig1}
    \vspace{-12pt}
\end{figure*}

However, most compensation-based methods share a common limitation: the quantization objective is characterized solely through the input activation covariance $\mathbf{H}_X$, which captures only the input-side second-order information of the weight space. Under the Kronecker-factored approximation (K-FAC)~\citep{martens2015optimizing}, the full weight Hessian factorizes as $\mathbf{H} \approx \mathbf{H}_X \otimes \mathbf{H}_G$, where $\mathbf{H}_G = \mathbb{E}[\mathbf{gg}^\top]$ is the gradient covariance with $g$ as the output gradient. Yet compensation-based methods built on the GPTQ-style~\citep{frantar2022gptq} column-wise OBS solver leave $\mathbf{H}_G$ unused, implicitly treating all output directions as equally important by assuming $\mathbf{H}_G = \mathbf{I}$. As shown in Figure~\ref{fig1}(a), output channels vary substantially in gradient magnitude, making this a suboptimal approximation.


We propose \KronQ, a post-training quantization method built on the Kronecker-factored Hessian approximation $\mathbf{H} \approx \mathbf{H}_X \otimes \mathbf{H}_G$. By incorporating the gradient covariance $\mathbf{H}_G$ into the quantization objective, \KronQ{} accounts for output-side Hessian information, a dimension largely overlooked by existing compensation-based methods. Our contributions are as follows.
\begin{enumerate}
    \item \textbf{Kronecker-Factored Quantization Error.} We extend the layer-wise quantization objective to incorporate $\mathbf{H}_G$ under the K-FAC approximation and introduce bidirectional incoherence processing. Notably, $\mathbf{H}_G$ cancels in the quantization update, preserving the efficiency of the base quantizer.
    
    \item \textbf{Inter-Layer Mixed Precision via Joint Hessian Traces.}
    We derive a sublayer sensitivity metric from $\mathrm{tr}(\mathbf{H}_G) \cdot \mathrm{tr}(\mathbf{H}_X)$ that differentiates sublayers sharing identical input statistics, enabling principled bit-width allocation across sublayers.
    \item \textbf{Empirical Validation.}
    We evaluate \KronQ{} on LLaMA-2 and LLaMA-3 from 7B to 70B across weight-only and weight-and-activation settings at W2/W3/W4, achieving consistent state-of-the-art results with the largest gains at 2-bit, as shown in Figure~\ref{fig1}(b).
\end{enumerate}
\vspace{-5mm}
\section{Related Work}
\label{sec:related}
\vspace{-3mm}
\paragraph{Compensation-Based Post-Training Quantization.}
The Optimal Brain Surgeon (OBS) framework~\citep{hassibi1992second, lecun1989optimal} provides the theoretical basis for Hessian-guided weight perturbation. GPTQ~\citep{frantar2022gptq} scales this to LLMs by performing column-wise OBS updates using the input activation covariance $\mathbf{H}_X$ as a proxy Hessian, establishing the de facto standard for one-shot LLM quantization. BoA~\citep{kim2024boa} extends GPTQ with attention-aware Hessians that capture inter-layer interactions within the attention module. GPTAQ~\citep{li2025gptaq} improves upon GPTQ by correcting for input drift caused by prior-layer quantization errors that accumulate during sequential layer-wise quantization, substantially reducing perplexity at ultra-low bit-widths.
Two recent methods go beyond input-only statistics. GuidedQuant~\citep{kim2025guidedquant} reweights $\mathbf{H}_X$ with per-output-channel end-loss saliency under a block-diagonal Fisher approximation.
YAQA~\citep{tseng2025model}, the work most closely related to ours, estimates the Fisher of the KL divergence to the original model through two Kronecker-factored sketches: Sketch A power-iterates under a token-independence assumption, while Sketch B is an unbiased single-pass sketch that retains cross-token structure. Both feed a two-sided LDLQ solver. \KronQ{} targets the same curvature but factorizes it under the K-FAC assumptions, and leaves the rounding step exactly as GPTAQ, using $\mathbf{H}_G$ only for incoherence processing and mixed-precision allocation.



\vspace{-2mm}
\paragraph{Rotation-Based Post-Training Quantization.} QuaRot~\citep{ashkboos2024quarot} and SpinQuant~\citep{liu2024spinquant} apply orthogonal transforms to weight matrices and activations to suppress outliers, enabling low-bit weight-and-activation quantization. QuIP~\citep{chee2023quip} establishes the theoretical foundation, showing that incoherent weight and Hessian matrices yield lower quantization error, and QuIP\#~\citep{tseng2024quip} extends this with the randomized Hadamard transform and lattice-based vector quantization for practical 2-bit quantization. More recent methods explore learnable transformations, including Kronecker-structured affine maps~\citep{sun2024flatquant} and jointly optimized orthogonal and scaling transforms~\citep{hu2025ostquant}.
 

\textbf{Mixed-Precision Quantization.} Mixed-precision methods differ in granularity. Intra-layer methods allocate bit-widths within a layer: SliM-LLM~\citep{huang2024slim} uses weight saliency and CMPQ~\citep{chen2024channel} uses activation norms. Inter-layer methods assign one bit-width per sublayer: HAWQ-V2~\citep{dong2020hawq} ranks layers by Hessian traces, while AMQ~\citep{lee2025amq}, Q-Palette~\citep{lee2026q}, and HIGGS~\citep{malinovskii2025higgs} rely on evolutionary search, a multiple-choice knapsack formulation, and a linearity theorem, respectively. 
Differentiating Q, K, and V, which share $\mathbf{H}_X$, requires costly search or per-sublayer loss estimation in these methods, whereas the \KronQ{} score $\mathrm{tr}(\mathbf{H}_G)\cdot\mathrm{tr}(\mathbf{H}_X)$ breaks this degeneracy analytically at the cost of a single backward pass.


\vspace{-3mm}
\section{Preliminary}
\label{sec:background}
\vspace{-3mm}
\subsection{Compensation-based Quantization}
\label{sec:bg:ptq}
Let $\mathbf{X} \in \mathbb{R}^{d_{\mathrm{in}} \times n}$ be a matrix of $n$ calibration inputs stacked as columns, $\mathbf{W} \in \mathbb{R}^{d_{\mathrm{out}} \times d_{\mathrm{in}}}$ the weight matrix of a linear layer with input dimension $d_{\mathrm{in}}$ and output dimension $d_{\mathrm{out}}$, and $\widehat{\mathbf{W}}$ its quantized counterpart. The standard layer-wise PTQ objective minimizes the squared output reconstruction error:
\begin{equation}
    \min_{\widehat{\mathbf{W}}} \; \|\mathbf{W}\mathbf{X} - \widehat{\mathbf{W}}\mathbf{X}\|_F^2
    \;=\;
    \mathrm{tr}\!\left[(\mathbf{W} - \widehat{\mathbf{W}})\, \mathbf{H}_X\, (\mathbf{W} - \widehat{\mathbf{W}})^\top\right],
    \label{eq:ptq_obj}
\end{equation}
where $\mathbf{H}_X = \mathbf{X}\mathbf{X}^\top \in \mathbb{R}^{d_{\mathrm{in}} \times d_{\mathrm{in}}}$ is the input activation covariance, which serves as a proxy for the layer-wise Hessian. GPTQ~\citep{frantar2022gptq} solves this objective greedily column by column using the OBS~\citep{hassibi1992second} update. The compensation applied to the remaining unquantized columns after quantizing column $p$ is given by
\begin{equation}
    \Delta\mathbf{W}_{:,p+1:}
    = -\frac{\mathbf{W}_{:,p} - \widehat{\mathbf{W}}_{:,p}}{[\mathbf{H}_X^{-1}]_{pp}}
      \cdot [\mathbf{H}_X^{-1}]_{p,p+1:},
    \label{eq:obs_update}
\end{equation}
where $[\mathbf{H}_X^{-1}]_{pp}$ is the $p$-th diagonal entry of the inverse Hessian and $[\mathbf{H}_X^{-1}]_{p,p+1:}$ is its $p$-th row restricted to columns $p+1$ onward. This distributes the quantization error of $\mathbf{W}_{:,p}$ optimally to the remaining weights. However, the layer-wise framework causes quantization errors to accumulate across layers, as the objective in Equation~\eqref{eq:ptq_obj} is evaluated on activations already degraded by prior-layer quantization. GPTAQ~\citep{li2025gptaq} addresses this by correcting for input drift via asymmetric calibration:
\begin{equation}
    \min_{\widehat{\mathbf{W}}} \; \|\mathbf{W}\widetilde{\mathbf{X}} - 
    \widehat{\mathbf{W}}\mathbf{X}\|_F^2
    \;=\;
    \min_{\widehat{\mathbf{W}}} \mathrm{tr}\!\left[
\Delta\mathbf{W}\,\mathbf{H}_X\,\Delta\mathbf{W}^\top 
+2
\mathbf{W}\,\Delta\mathbf{X}\mathbf{X}^\top\,\Delta\mathbf{W}^\top
\right],
    \label{eq:gptaq_obj}
\end{equation}
where $\Delta\mathbf{W} = \mathbf{W} - \widehat{\mathbf{W}}$, $\mathbf{X}$ is the input from the running quantized model, $\widetilde{\mathbf{X}}$ is the corresponding activation of the full-precision model, and $\Delta\mathbf{X} = \widetilde{\mathbf{X}} - \mathbf{X}$ is the input drift. The term $\|\mathbf{W}\Delta\mathbf{X}\|_F^2$ is constant in $\widehat{\mathbf{W}}$ and omitted. Despite this correction, GPTAQ inherits the same proxy Hessian $\mathbf{H}_X$, leaving the output-side gradient statistics unaccounted for.

\subsection{Incoherence Processing}
\label{sec:bg:incoherence}
Incoherence processing~\citep{chee2023quip} is a preprocessing technique that reduces quantization error by spreading weight magnitudes uniformly across all dimensions. Formally, the incoherence of $\mathbf{W}$ and its proxy Hessian $\mathbf{H}_X$ are measured by
\begin{equation}
    \mu(\mathbf{W}) = \frac{\sqrt{d_{\mathrm{out}} d_{\mathrm{in}}}}{\|\mathbf{W}\|_F}
    \max_{i,j} |\textbf{W}_{ij}|,
    \qquad
    \mu(\mathbf{H}_X) = \sqrt{d_{\mathrm{in}}} \cdot \max_{i,j} |\textbf{Q}_{ij}|,
    \label{eq:incoherence_measure}
\end{equation}
where $\mathbf{Q}$ denotes the eigenvector matrix of $\mathbf{H}_X$, and lower $\mu$ leads to tighter quantization error bounds. Intuitively, this is achieved by randomizing the spectral directions of $\mathbf{W}$ and $\mathbf{H}_X$, so that rounding sensitivity is no longer concentrated along specific coordinate axes. To this end, random orthogonal matrices $\mathbf{U} \in \mathbb{R}^{d_{\mathrm{out}} \times d_{\mathrm{out}}}$ and $\mathbf{V} \in \mathbb{R}^{d_{\mathrm{in}} \times d_{\mathrm{in}}}$ are applied:
\begin{equation}
    \mathbf{W} \leftarrow \mathbf{U}\mathbf{W}\mathbf{V}^\top,
    \quad \mathbf{H}_X \leftarrow \mathbf{V} \mathbf{H}_X \mathbf{V}^\top.
    \label{eq:incoherence}
\end{equation}
After quantization, the rotations must be applied online during inference. QuIP~\citep{chee2023quip} instantiates $\mathbf{U}$ and $\mathbf{V}$ as Kronecker-structured random orthogonal matrices, introducing $\Theta(d_{\mathrm{in}}^{3/2} + d_{\mathrm{out}}^{3/2})$ inference overhead per layer. QuIP\#~\citep{tseng2024quip} replaces these with randomized Hadamard transforms, reducing this to $\Theta(d_{\mathrm{in}} \log d_{\mathrm{in}} +  d_{\mathrm{out}} \log d_{\mathrm{out}})$. However, this incoherence process accounts only for input-side Hessian information through $\mathbf{H}_X$, leaving the output-side entirely unaddressed.
\vspace{-2mm}
\section{Method}
\label{sec:method}

\vspace{-2mm}
\subsection{Kronecker-Factored Quantization Error}
\label{sec:method:kfac}

The standard layer-wise PTQ objective in Equation~\eqref{eq:ptq_obj} uses $\mathbf{H}_X$ as a proxy for the full Hessian $\mathbf{H}$, discarding the output-side statistics entirely. We recover this missing factor via the Kronecker-factored approximation~\citep{martens2015optimizing,van2023llm}. For a linear layer $\mathbf{y} = \mathbf{W}\mathbf{x}$, the per-sample gradient factorizes as $\frac{\partial \mathcal{L}}{\partial \mathbf{W}} = \mathbf{g}\mathbf{x}^\top$, where $\mathbf{g} = \partial\mathcal{L}/\partial\mathbf{y}$. 
The true Fisher approximation of $\mathbf{H}$, which samples labels from the model's own output distribution rather than using the empirical Fisher~\citep{kunstner2019limitations}, gives:
\begin{align}
    \mathbf{H}
    &\;=\; \mathbb{E}_{\mathbf{x},\, y \sim p(\cdot \mid \mathbf{x})}\!\left[
        \mathrm{vec}\!\left(\frac{\partial \mathcal{L}}{\partial \mathbf{W}}\right)
        \mathrm{vec}\!\left(\frac{\partial \mathcal{L}}{\partial \mathbf{W}}\right)^\top
    \right]
    \;=\; \mathbb{E}\!\left[
        (\mathbf{x} \otimes \mathbf{g})(\mathbf{x} \otimes \mathbf{g})^\top
    \right]
    \;=\; \mathbb{E}\!\left[
        \mathbf{x}\mathbf{x}^\top \otimes \mathbf{g}\mathbf{g}^\top
    \right].
    \label{eq:fisher_exact}
\end{align}
Since the joint expectation $\mathbb{E}[\mathbf{x}\mathbf{x}^\top \otimes \mathbf{g}\mathbf{g}^\top] \in \mathbb{R}^{d_{\mathrm{out}}d_{\mathrm{in}} \times d_{\mathrm{out}}d_{\mathrm{in}}}$ is intractable, we apply the K-FAC independence assumption $\mathbf{x} \perp\!\!\!\perp \mathbf{g}$~\citep{martens2015optimizing,botev2017practical}, under which the expectation factorizes:
\begin{equation}
    \mathbf{H} \;\approx\; \mathbb{E}[\mathbf{x}\mathbf{x}^\top] \otimes
    \mathbb{E}[\mathbf{g}\mathbf{g}^\top]
    \;=:\; \mathbf{H}_X \otimes \mathbf{H}_G,
    \label{eq:kfac}
\end{equation}
reducing complexity from $\mathcal{O}(d_{\mathrm{in}}^2 d_{\mathrm{out}}^2)$ to $\mathcal{O}(d_{\mathrm{in}}^2 + d_{\mathrm{out}}^2)$. Substituting \eqref{eq:kfac} into \eqref{eq:ptq_obj} yields the Kronecker-factored quantization objective:
\begin{equation}
    \min_{\widehat{\mathbf{W}}} \;
    \mathrm{tr}\!\left[\mathbf{H}_G\,\Delta\mathbf{W}\,\mathbf{H}_X\,
    \Delta\mathbf{W}^\top\right].
    \label{eq:kron_obj}
\end{equation}
Most existing PTQ methods implicitly set $\mathbf{H}_G = \mathbf{I}$. However, as shown in Figure~\ref{fig1}(a), output channels vary substantially in gradient magnitude, making this approximation suboptimal. \KronQ{} retains $\mathbf{H}_G$, estimated via a single backward pass over the calibration set and stored as a $d_{\mathrm{out}} \times d_{\mathrm{out}}$ matrix. Building on GPTAQ~\citep{li2025gptaq}, we further correct for input drift by incorporating the asymmetric correction from Equation~\eqref{eq:gptaq_obj}:
\begin{equation}
    \min_{\widehat{\mathbf{W}}} \mathrm{tr}\!\left[\mathbf{H}_G\!\left(
\Delta\mathbf{W}\,\mathbf{H}_X\,\Delta\mathbf{W}^\top 
+2 \mathbf{W}\,\Delta\mathbf{X}\mathbf{X}^\top\,\Delta\mathbf{W}^\top
\right)\right].
    \label{eq:kronq_full}
\end{equation}
The optimal weight update under \eqref{eq:kronq_full} is characterized in the following proposition.

\begin{proposition}[\KronQ{} Weight Compensation]
\label{prop:hg_cancel}
Under the column-wise OBS update applied to \eqref{eq:kronq_full}, $\mathbf{H}_G$ cancels algebraically, and the weight compensation after quantizing column $p$ reduces to:
\begin{equation}
    \Delta\mathbf{W}_{:,p+1:}
    = -\boldsymbol{\delta}_p\, [\mathbf{H}_X^{-1}]_{p,p+1:}
    + \mathbf{W}_{:,p} \cdot [\mathbf{P}]_{p,p+1:},
    \label{eq:kronq_update}
\end{equation}
where $\boldsymbol{\delta}_p = (\mathbf{W}_{:,p} - \widehat{\mathbf{W}}_{:,p})/[\mathbf{H}_X^{-1}]_{pp}$ 
is the scaled quantization error, 
$\mathbf{P} = \alpha\left((\Delta\mathbf{X}\mathbf{X}^\top \mathbf{L}) \odot \mathbf{M}_U\right)\mathbf{L}^\top$ 
the GPTAQ asymmetric correction with $\mathbf{H}_X^{-1} = \mathbf{L}\mathbf{L}^\top$, 
and $\mathbf{M}_U$ the strictly upper-triangular mask.
\end{proposition}
The derivation of Proposition~\ref{prop:hg_cancel} is provided in Appendix~\ref{proof_prop1}. \KronQ{} inherits the computational efficiency of GPTAQ while exploiting $\mathbf{H}_G$ for bidirectional incoherence processing and mixed-precision allocation.
 

\begin{figure*}
    \centering
    \includegraphics[width=\textwidth]{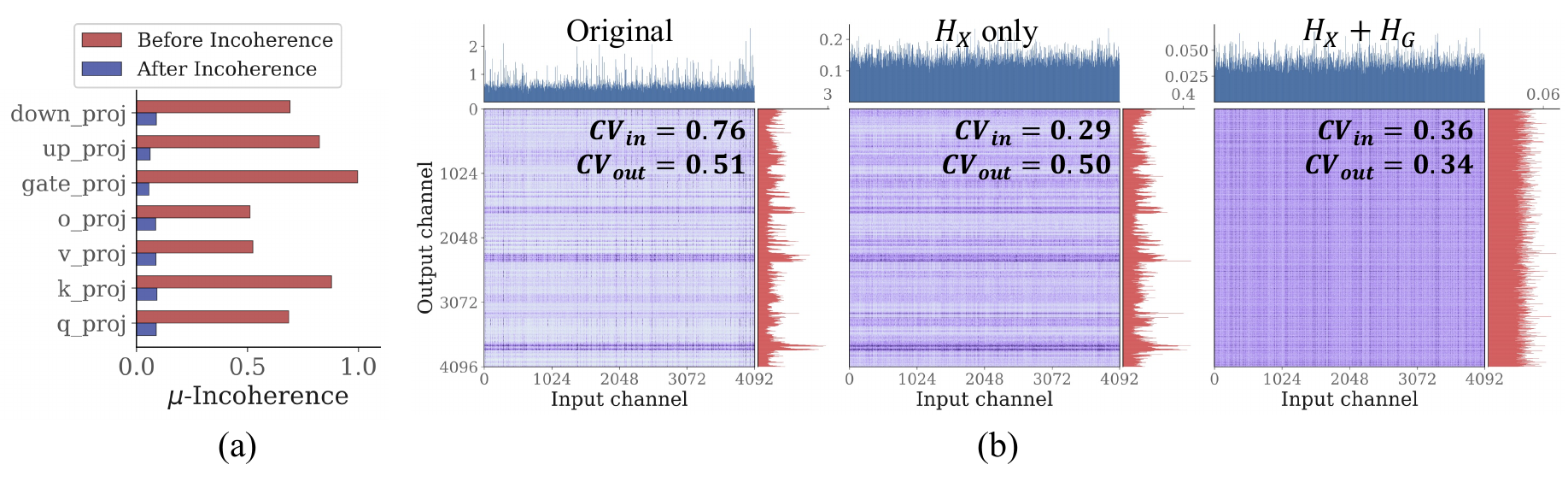}
    \vspace{-8mm}
    \caption{(a) $\mu$-incoherence of $\mathbf{H}_G$ before and after incoherence preprocessing across sublayers of LLaMA-2-7B.  (b) Weight magnitude distribution of \texttt{Q\_proj} under three configurations: original weights, after input-side incoherence ($\mathbf{H}_X$ only), and after bidirectional incoherence ($\mathbf{H}_X + \mathbf{H}_G$), where $\mathrm{CV}_{\mathrm{in}}$ and $\mathrm{CV}_{\mathrm{out}}$ denote the coefficient of variation of column and row norms, respectively.}
    \label{fig2}
    \vspace{-14pt}
\end{figure*}

\vspace{-3mm}
\subsection{Bidirectional Incoherence Processing}
\vspace{-2mm}
\label{sec:method:biip}

\citet{chee2023quip} and \citet{tseng2024quip} show that quantization error can be reduced when the proxy Hessian is incoherent---its eigenvectors are not aligned with the coordinate axes---and that $\mathbf{H}_X$ is highly coherent in practice, motivating column-side incoherence processing. Under the \KronQ{} objective \eqref{eq:kronq_full}, $\mathbf{H}_G$ appears as the complementary output-side Hessian factor. This raises a natural question: \textbf{\emph{is $\mathbf{H}_G$ also coherent?}}

We answer this via the incoherence measure $\mu(\mathbf{H}_G) = \sqrt{d_{\mathrm{out}}} \cdot \max_{i,j} |\textbf{Q}_{ij}|$ from Equation~\eqref{eq:incoherence_measure}. Here, $\mu/\sqrt{d_{\mathrm{out}}} \to 1$ indicates maximal coherence and $\mu/\sqrt{d_{\mathrm{out}}} \to 0$ indicates incoherence. Figure~\ref{fig2}(a) presents $\mu(\mathbf{H}_G)/\sqrt{d_{\mathrm{out}}}$ before and after incoherence processing across sublayers of LLaMA-2-7B. $\mu(\mathbf{H}_G)/\sqrt{d_{\mathrm{out}}}$ reaches up to $0.99$, but it drops below $0.10$ for all sublayers after incoherence processing, confirming that output-side rotation effectively renders $\mathbf{H}_G$ incoherent.
This motivates bidirectional incoherence processing (BiIP). Extending the input-side diagonal rescaling of~\citet{chee2023quip} to both column and row directions, we apply:
\begin{equation}
    \mathbf{W} \;\leftarrow\; \mathbf{S}_G\, \mathbf{W}\, \mathbf{S}_X,
\qquad
\mathbf{S}_X = \mathrm{diag}\!\left(\frac{[\mathbf{H}_X]_{jj}}
{\|\mathbf{W}_{:,j}\|^2}\right)^{1/4}\!,
\quad
\mathbf{S}_G = \mathrm{diag}\!\left(\frac{[\mathbf{H}_G]_{ii}}
{\|\mathbf{W}_{i,:}\|^2}\right)^{1/4}\!,
    \label{eq:rescale}
\end{equation}
where the output-side $\mathbf{S}_G$ term, derived from $\mathbf{H}_G$, is 
novel. Since the subsequent quantization operates in the rescaled weight space, the Hessians and drift statistics are transformed as $\mathbf{H}_X \leftarrow \mathbf{S}_X^{-1} \mathbf{H}_X \mathbf{S}_X^{-1}$, $\mathbf{H}_G \leftarrow \mathbf{S}_G^{-1} \mathbf{H}_G \mathbf{S}_G^{-1}$, $\Delta\mathbf{X}\mathbf{X}^\top \leftarrow \mathbf{S}_X^{-1} \Delta\mathbf{X}\mathbf{X}^\top \mathbf{S}_X^{-1}$, leaving the objective~\eqref{eq:kronq_full} unchanged. We then apply orthogonal transforms:
\begin{equation}
    \mathbf{W} \leftarrow \mathbf{U} \mathbf{W} \mathbf{V}^\top,
    \quad
    \mathbf{H}_X \leftarrow \mathbf{V} \mathbf{H}_X \mathbf{V}^\top,
    \quad
    \mathbf{H}_G \leftarrow \mathbf{U} \mathbf{H}_G \mathbf{U}^\top,
    \quad
    \Delta\mathbf{X}\mathbf{X}^\top \leftarrow \mathbf{V} \Delta\mathbf{X}\mathbf{X}^\top \mathbf{V}^\top,
    \label{eq:biip}
\end{equation}
instantiating $\mathbf{U}$ and $\mathbf{V}$ as randomized Hadamard transforms~\citep{tseng2024quip} to make both $\mathbf{H}_G$ and $\mathbf{H}_X$ incoherent with high probability. Figure~\ref{fig2}(b) shows that incoherencing $\mathbf{H}_X$ alone leaves the output-channel coefficient of variation ($CV_{out}$) nearly unchanged, whereas bidirectional incoherence processing reduces both $CV_{in}$ to $0.36$ and $CV_{out}$ to $0.34$. At inference, BiIP introduces no additional overhead beyond QuIP\#~\citep{tseng2024quip}. The diagonal rescaling $\mathbf{S}_X$ and $\mathbf{S}_G$ are reverted elementwise with negligible cost, while $\mathbf{U}$ and $\mathbf{V}$ introduce $\Theta(d_{\mathrm{in}} \log d_{\mathrm{in}} + d_{\mathrm{out}} \log d_{\mathrm{out}})$ per layer. The \KronQ{} objective stays invariant under these transformations, as formalized in Theorem~\ref{thm:invariance} with proof in Appendix~\ref{proof:theorem1}.



\begin{theorem}
\label{thm:invariance}
For any orthogonal $\mathbf{U} \in \mathbb{R}^{d_{\mathrm{out}}\times d_{\mathrm{out}}}$ and $\mathbf{V} \in \mathbb{R}^{d_{\mathrm{in}}\times d_{\mathrm{in}}}$,  \KronQ{} objective $\mathrm{tr}\!\left[\mathbf{H}_G\!\left(
\Delta\mathbf{W}\,\mathbf{H}_X\,\Delta\mathbf{W}^\top 
+2 \mathbf{W}\,\Delta\mathbf{X}\mathbf{X}^\top\,\Delta\mathbf{W}^\top
\right)\right]$ is invariant under the transformation \eqref{eq:biip}.
\end{theorem}


\citet{tseng2025model} bound the two-sided LDLQ proxy loss by the product of the LDL diagonal traces of both factors, where $\mathbf{H}_G$ enters the solver as additional feedback. In our case, $\mathbf{H}_G$ cancels in the column-wise update, leaving a one-sided solver. The following results show that $\mathrm{diag}(\mathbf{H}_G)$ suffices, and that $\mathbf{S}_G$ in Equation~\eqref{eq:rescale} is optimal to leading order.

\begin{proposition}[BiIP-Preconditioned Expected Loss]\label{prop:biip}
Let $h_i=[\mathbf{H}_G]_{ii}$, $a_i=\lVert\mathbf{W}_{i,:}\rVert^{2}$, $\mathbf{D}$ the LDL diagonal of the transformed $\mathbf{H}_X$, and $L_q$ the objective of Equation~\eqref{eq:kron_obj}. Among diagonal output rescalings applied before the rotation,
\begin{equation}
\mathbb{E}[L_q] \;\propto\; \Big(\sum_{i}\frac{h_i}{s_i^{2}}\Big)\Big(\sum_{j}s_j^{2}a_j\Big)\operatorname{tr}(\mathbf{D}),
\label{eq:biip-bound}
\end{equation}
minimized at $s_i=(h_i/a_i)^{1/4}$, the $\mathbf{S}_G$ of Equation~\eqref{eq:rescale}.
\end{proposition}
This improves over input-side-only processing by $\rho=\big(\sum_i\sqrt{h_i a_i}\big)^{2}\big/\big(\sum_i h_i\sum_i a_i\big)\le1$, strict unless $h_i\propto a_i$, which is far from the case in practice as $\operatorname{diag}(\mathbf{H}_G)$ spans orders of magnitude (Figure~\ref{fig1}(a)). The full derivation and the relation to \citet{tseng2025model} are given in Appendix~\ref{proof:prop2}. The overall algorithms are summarized in Algorithm~\ref{algo1} and~\ref{algo2}.

\begin{figure*}
    \centering
    \includegraphics[width=\textwidth]{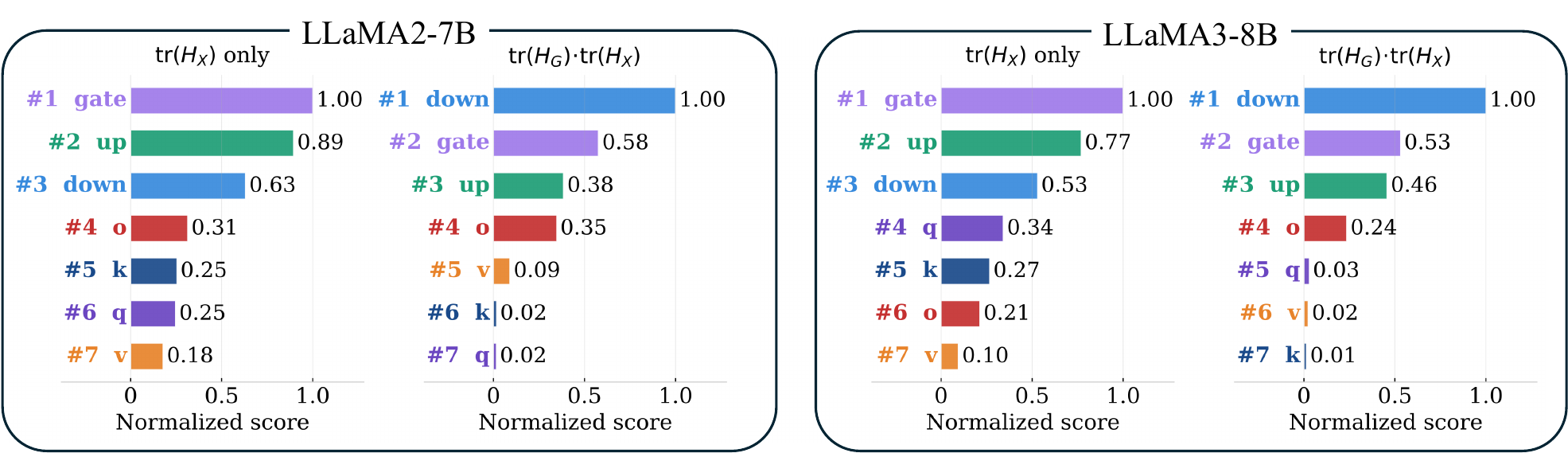}
    \vspace{-6mm}
    \caption{Sublayer sensitivity rankings under the \KronQ{} score
$\mathrm{tr}(\mathbf{H}_G)\cdot\mathrm{tr}(\mathbf{H}_X)$ and the activation-only score $\mathrm{tr}(\mathbf{H}_X)$ on LLaMA-2-7B and LLaMA-3-8B.}
    
    \label{fig3}
    \vspace{-12pt}
\end{figure*}

\vspace{-3mm}
\subsection{Mixed-Precision Allocation via Joint Hessian Traces}
\label{sec:method:mp}
\vspace{-2mm}
Mixed-precision quantization allocates a bit budget across sublayers by ranking them according to a sensitivity metric. Under the second-order approximation, the expected quantization loss for layer $\ell$ is proportional to $\epsilon_\ell^2 \cdot \mathrm{tr}(\mathbf{H}^{(\ell)})$~\citep{dong2020hawq}, where $\epsilon_\ell^2$ denotes the per-element rounding error variance at bit-width $b_\ell$. Since the goal is to rank sublayers rather than compute absolute loss values, $\epsilon_\ell^2$ is treated as a constant within each layer~\citep{dong2020hawq}, and sensitivity reduces to $\mathrm{tr}(\mathbf{H}^{(\ell)})$. Existing methods approximate $\mathbf{H}^{(\ell)} \approx \mathbf{H}_X^{(\ell)}$, yielding the activation-only score $\mathrm{tr}(\mathbf{H}_X^{(\ell)})$. Under the Kronecker factorization $\mathbf{H} \approx \mathbf{H}_X \otimes \mathbf{H}_G$, this extends naturally to both factors:
\begin{equation}
    \mathbb{E}[\mathcal{L}_\ell] \;\propto\;
    \mathrm{tr}(\mathbf{H}_G^{(\ell)}) \cdot \mathrm{tr}(\mathbf{H}_X^{(\ell)}).
    \label{eq:sensitivity}
\end{equation}
We define the \KronQ{} sensitivity score as $s_\ell = \mathrm{tr}(\mathbf{H}_G^{(\ell)}) \cdot \mathrm{tr}(\mathbf{H}_X^{(\ell)})$, computed post-BiIP, and allocate higher bit-widths to sublayers with larger $s_\ell$. Within a transformer attention block, the query (Q), key (K), and value (V) projections share the same input, so $\mathbf{H}_X^{(Q)} = \mathbf{H}_X^{(K)} = \mathbf{H}_X^{(V)}$: any activation-only metric assigns identical sensitivity to all three. The \KronQ{} score breaks this degeneracy via $\mathbf{H}_G$, which differs across Q, K, and V as they receive different downstream gradients. As shown in Figure~\ref{fig3}, the sublayer rankings under $\mathrm{tr}(\mathbf{H}_G)\cdot\mathrm{tr}(\mathbf{H}_X)$ differ substantially from those under $\mathrm{tr}(\mathbf{H}_X)$ alone. The \KronQ{} score yields strictly better perplexity--bit-width tradeoffs, as we quantify in Section~\ref{sec:exp:mp}.



\vspace{-4mm}
\section{Experiments}
\label{sec:experiments}
\vspace{-4mm}
\subsection{Experimental Setup}
\label{sec:exp:setup}
\vspace{-3mm}
We evaluate \KronQ{} on the LLaMA family, from LLaMA-2-7B/13B/70B~\citep{touvron2023llama} to LLaMA-3-8B/70B~\citep{grattafiori2024llama}. We compare against GPTQ~\citep{frantar2022gptq} and GPTAQ~\citep{li2025gptaq} as primary baselines. We consider three regimes: (i) weight-only quantization (WxA16), (ii) group quantization (WxA16, $g{=}128$), and (iii) weight-and-activation quantization (WxA4). For activation quantization, we utilize the QuaRot~\citep{ashkboos2024quarot} framework. All other quantization configurations follow GPTAQ~\citep{li2025gptaq}. Perplexity is measured on WikiText-2~\citep{merity2016pointer} and zero-shot accuracy is reported on seven commonsense reasoning benchmarks: PiQA~\citep{bisk2020piqa}, Arc-Easy (ArcE), Arc-Challenge (ArcC)~\citep{clark2018think}, HellaSwag (HS)~\citep{zellers2019hellaswag}, WinoGrande (WG)~\citep{sakaguchi2021winogrande}, BoolQ~\citep{clark2019boolq}, and OpenBookQA (OBQA)~\citep{mihaylov2018can}. All experiments use 128 calibration samples from WikiText-2 with 2048 context lengths and are run on A100 GPUs. $\mathbf{H}_G$ is precomputed via a single backward pass prior to calibration and loaded layer-by-layer on demand, incurring no additional overhead during quantization. Results for prior works are taken from their respective papers, except for OmniQuant~\citep{shao2023omniquant} zero-shot accuracy, which we reproduce using the official implementation.

\begin{table*}[]
    \caption{Weight-only quantization (WxA16). WikiText-2 perplexity $\downarrow$ across LLaMA-2/3 models, and zero-shot accuracy $\uparrow$ for LLaMA-2-7B on three reasoning benchmarks (PiQA, ArcC, WG). NaN indicates diverged quantization, and {--} indicates results not available. Avg is reported only when all three tasks are available.}
    \vspace{-3mm}
    \label{tab:weight_only}
    \resizebox{\linewidth}{!}{%
    \begin{NiceTabular}{c||l|ccc|cc|cccc}
\toprule
\textbf{Bits} & \textbf{Method}
  & \textbf{L2-7B} & \textbf{L2-13B} & \textbf{L2-70B}
  & \textbf{L3-8B} & \textbf{L3-70B}
  & \textbf{PiQA} & \textbf{ArcC} & \textbf{WG} & \textbf{Avg} \\
\midrule
FP16 & --   & 5.47  & 4.88  & 3.32  & 6.14  & 2.85  & 78.5 & 40.6 & 67.3 & 62.1 \\
\midrule
\multirow{9}{*}{4}
 & OmniQuant & 5.74  & 5.02  & 3.47  & --    & --    & 77.5 & 39.1 & 67.0 & 61.2 \\
 & AWQ       & 5.83  & 5.06  & 3.48  & 7.11  & --    & 76.0   & --   & --   & --   \\
  & BoA       & 5.62  & 5.10  & --    & 6.56  & --    & 78.4 & 43.9 & 69.1 & 63.8 \\
  & SpinQuant       & 5.58  & 5.00  & 3.43  & 6.49  & 3.49    & 77.9   & 43.3   & 67.5   &  62.9  \\
  & OSTQuant       & 5.64  & 4.94  & 3.41  & 6.53  & 3.19    & 78.9   & 44.5   & 68.6   &  64.0  \\
 & QEP       & 5.75  & 5.03  & 3.49  & 6.65  & --    & 76.3   & --   & --   & --   \\
 & QuIP      & 8.43  & 5.14  & 3.83  & 7.00  & --    & --   & --   & --   & --   \\
 & QuIP\#    & 5.66  & 5.00  & 3.42  & --    & --    & 77.2 & 40.4 & 67.5 & 61.7 \\
 & GPTQ      & 5.82  & 5.37  & 3.50  & 8.61  & 27.49 & 77.2 & 41.2 & 68.7 & 62.4 \\
 & GPTAQ     & 5.69  & 5.08  & 3.46  & 6.80  & 399.46& 76.8 & 41.6 & 68.4 & 62.3 \\
 & \cellcolor{gray!15}\textbf{\KronQ}
             & \cellcolor{gray!15}\textbf{5.56}
             & \cellcolor{gray!15}\textbf{4.95}
             & \cellcolor{gray!15}\textbf{3.40}
             & \cellcolor{gray!15}\textbf{6.42}
             & \cellcolor{gray!15}\textbf{3.25}
             & \cellcolor{gray!15}\textbf{78.7}
             & \cellcolor{gray!15}\textbf{44.9}
             & \cellcolor{gray!15}\textbf{68.4}
             & \cellcolor{gray!15}\textbf{64.0} \\
\midrule
\multirow{9}{*}{3}
 & OmniQuant & 6.58  & 5.58  & 3.92  & --    & --    & 74.0 & 36.3 & 63.6 & 58.0 \\
 & AWQ       & 15.30 & 6.45  & 4.36  & 11.80 & --    & 64.7   & --   & --   & --   \\
 & BoA       & 6.01  & 5.83  & --    & 7.78  & --    & 78.4 & 40.3 & 67.6 & 62.1 \\
 & QEP       & 6.15  & 5.35  & 3.81  & 7.70  & --    & 72.5  & --   & --   & --   \\
 & QuIP      & 12.05 & 5.50  & 4.14  & 8.29  & --    & --   & --   & --   & --   \\
 & QuIP\#    & 6.19  & 5.34  & 3.71  & --    & --    & 76.1 & 38.1 & 65.1 & 59.8 \\

 & GPTQ      & 6.74  & 9.41  & 5.47  & 8.24  & 2.6e3 & 76.0 & 36.5 & 65.7 & 59.4 \\
 & GPTAQ     & 6.21  & 6.52  & 3.93  & 7.84  & 1.6e4 & 75.4 & 37.2 & 65.3 & 59.3 \\
 & \cellcolor{gray!15}\textbf{\KronQ}
             & \cellcolor{gray!15}\textbf{5.83}
             & \cellcolor{gray!15}\textbf{5.18}
             & \cellcolor{gray!15}\textbf{3.66}
             & \cellcolor{gray!15}\textbf{7.09}
             & \cellcolor{gray!15}\textbf{4.41}
             & \cellcolor{gray!15}\textbf{76.9}
             & \cellcolor{gray!15}\textbf{43.3}
             & \cellcolor{gray!15}\textbf{67.2}
             & \cellcolor{gray!15}\textbf{62.5} \\
\midrule
\multirow{9}{*}{2}
 & OmniQuant & 37.73 & 17.22 & 7.81  & --    & --    & 57.2 & 26.1 & 51.5 & 44.9 \\
 & AWQ       & NaN   & NaN   & NaN   & NaN   & --    & 50.7   & --   & --   & --   \\
 & BoA       & 12.76 & 18.33 & --    & 21.70 & --    & 64.9 & 28.3 & 54.5 & 49.2 \\
 & QEP       & 11.97 & 8.42  & 5.87  & 27.33 & --    & 50.5  & --   & --   & --   \\
 & QuIP      & 65.59 & 11.23 & 6.54  & 70.52 & --    & 54.6 & 19.4 & 51.8 & 41.9 \\
 & QuIP\#    & 12.30 & 7.60  & 4.87  & --    & --    & 68.0 & 29.2 & 59.0 & 52.1 \\

 & GPTQ      & 31.11 & 35.89 & 9.54  & 25.43 & 2.6e3 & 63.0 & 28.4 & 53.9 & 48.4 \\
 & GPTAQ     & NaN   & 20.43 & 6.41  & 18.27 & NaN   & NaN  & NaN  & NaN  & NaN  \\
 & \cellcolor{gray!15}\textbf{\KronQ}
             & \cellcolor{gray!15}\textbf{8.19}
             & \cellcolor{gray!15}\textbf{6.99}
             & \cellcolor{gray!15}\textbf{5.14}
             & \cellcolor{gray!15}\textbf{11.92}
             & \cellcolor{gray!15}\textbf{7.93}
             & \cellcolor{gray!15}\textbf{68.0}
             & \cellcolor{gray!15}\textbf{31.3}
             & \cellcolor{gray!15}\textbf{60.8}
             & \cellcolor{gray!15}\textbf{53.3} \\
\bottomrule
\end{NiceTabular}}
    \vspace{-14pt}
\end{table*}

\begin{table}[t]
\centering
\begin{minipage}[t]{0.44\linewidth}
\centering
\caption{Comparison with GuidedQuant (LLaMA-2, WikiText-2, ctx 4096).}
\label{tab:guidedquant}
\vspace{-2pt}
\scalebox{0.8}{%
\begin{NiceTabular}{llcc}[code-before = \rectanglecolor{gray!15}{2-4}{7-4}]
\toprule
Model & Bits & LNQ+GuidedQuant & \textbf{\KronQ{}} \\
\midrule
L2-7B  & W3 & 5.57 & \textbf{5.45} \\
L2-13B & W3 & 4.91 & \textbf{4.84} \\
L2-70B & W3 & 3.47 & \textbf{3.48} \\
\midrule
L2-7B  & W2 & 8.83 & \textbf{7.60} \\
L2-13B & W2 & 7.26 & \textbf{6.50} \\
L2-70B & W2 & 5.04 & \textbf{5.07} \\
\bottomrule
\end{NiceTabular}%
}
\end{minipage}\hfill
\begin{minipage}[t]{0.52\linewidth}
\centering
\caption{Comparison with YAQA (WikiText-2 PPL, identical INT per-channel grids,
no finetuning, L2-7B ctx 2048, L3.1-8B ctx 8192).}
\label{tab:yaqa}
\vspace{-4pt}
\scalebox{0.8}{%
\begin{tabular}{lcccc}
\toprule
& \multicolumn{2}{c}{L2-7B} & \multicolumn{2}{c}{L3.1-8B} \\
\cmidrule(lr){2-3}\cmidrule(lr){4-5}
Method & W2 & W4 & W2 & W4 \\
\midrule
YAQA-B (RedPajama) & 10.20 & 5.57 & 18.27 & 6.71 \\
YAQA-B (WikiText2) & 10.07 & 5.57 & 19.42 & 6.73 \\
\KronQ{}+GPTQ         & 10.18 & 5.58 & 14.32 & 6.72 \\
\cellcolor{gray!15}\textbf{\KronQ{}+GPTAQ} & \cellcolor{gray!15}\textbf{8.19} & \cellcolor{gray!15}\textbf{5.56} & \cellcolor{gray!15}\textbf{11.82} & \cellcolor{gray!15}\textbf{6.69} \\
\bottomrule
\end{tabular}%
}
\end{minipage}
\vspace{-1.5mm}
\end{table}

\begin{table}[t]
    \caption{WikiText-2 perplexity and zero-shot accuracy for weight-only group quantization ($g{=}128$) at W2.}
    \vspace{-2mm}
    \label{tab:group_w2}
    \resizebox{\linewidth}{!}{%
    \begin{NiceTabular}{l||l|c|c|cccccccr}
\toprule
\textbf{Model} & \textbf{Method} & \textbf{Bits}
  & \textbf{Wiki2} $\!\downarrow$
  & \textbf{PiQA} & \textbf{ArcE} & \textbf{ArcC} & \textbf{HS}
  & \textbf{WG} & \textbf{BoolQ} & \textbf{OBQA}
  & \textbf{Avg} $\!\uparrow$ \\
\midrule
\multirow{4}{*}{LLaMA-2-7B}
& OmniQuant  & 2 & 11.06 & 65.8 & 45.0 & 28.0 & 49.3 & 54.6 & 61.2 & 34.2 & 48.3 \\
 & GPTQ  & 2 & 274.00 & 57.8 & 38.5 & 26.3 & 29.3 & 53.1 & 42.1 & 28.2 & 39.3 \\
 & GPTAQ & 2 &  23.19 & 59.3 & 39.7 & 26.3 & 31.0 & 55.7 & 41.0 & 29.0 & 40.3 \\
 & \cellcolor{gray!15}\textbf{\KronQ} & \cellcolor{gray!15}\textbf{2} & \cellcolor{gray!15}\textbf{7.60} & \cellcolor{gray!15}\textbf{70.6} & \cellcolor{gray!15}\textbf{57.7} & \cellcolor{gray!15}\textbf{32.9} & \cellcolor{gray!15}\textbf{56.5} & \cellcolor{gray!15}\textbf{62.1} & \cellcolor{gray!15}\textbf{65.4} & \cellcolor{gray!15}\textbf{37.4} & \cellcolor{gray!15}\textbf{54.6} \\
 \midrule
\multirow{4}{*}{LLaMA-2-13B}
& OmniQuant  & 2 & 8.26 & 69.0 & 57.1 & 32.7 & 56.5 & 53.1 & 63.7 & 35.6 & 52.5 \\
 & GPTQ  & 2 &   8.40 & 67.6 & 43.6 & 29.4 & 49.3 & 61.3 & 42.1 & 35.0 & 46.9 \\
 & GPTAQ & 2 &   7.17 & 65.1 & 46.4 & 29.5 & 47.6 & 60.2 & 57.0 & 32.8 & 48.4 \\
 & \cellcolor{gray!15}\textbf{\KronQ} & \cellcolor{gray!15}\textbf{2} & \cellcolor{gray!15}\textbf{6.51} & \cellcolor{gray!15}\textbf{72.6} & \cellcolor{gray!15}\textbf{68.9} & \cellcolor{gray!15}\textbf{38.1} & \cellcolor{gray!15}\textbf{63.9} & \cellcolor{gray!15}\textbf{65.9} & \cellcolor{gray!15}\textbf{66.0} & \cellcolor{gray!15}\textbf{36.8} & \cellcolor{gray!15}\textbf{58.9} \\
\midrule
\multirow{3}{*}{LLaMA-3-8B}
 & GPTQ  & 2 &  16.41 & 55.2 & 34.2 & 20.9 & 45.1 & 58.7 & 59.3 & 29.4 & 43.3 \\
 & GPTAQ & 2 &  13.08 & 59.5 & 37.9 & 23.0 & 50.2 & 60.1 & 60.7 & 31.6 & 46.1 \\
 & \cellcolor{gray!15}\textbf{\KronQ} & \cellcolor{gray!15}\textbf{2} & \cellcolor{gray!15}\textbf{10.76} & \cellcolor{gray!15}\textbf{69.0} & \cellcolor{gray!15}\textbf{56.3} & \cellcolor{gray!15}\textbf{34.4} & \cellcolor{gray!15}\textbf{53.3} & \cellcolor{gray!15}\textbf{63.3} & \cellcolor{gray!15}\textbf{65.9} & \cellcolor{gray!15}\textbf{34.2} & \cellcolor{gray!15}\textbf{53.8} \\

\bottomrule
\end{NiceTabular}}
 \vspace{-1.5mm}
\end{table}

\begin{table}[!t]
\caption{WikiText-2 perplexity and zero-shot accuracy for weight \& activation quantization.}
\vspace{-2mm}
\label{tab:wa_w2}
\resizebox{\linewidth}{!}{%
\begin{NiceTabular}{l||l|c|c|ccccccccr}
\toprule
\textbf{Model} & \textbf{Method} & \textbf{Bits}
  & \textbf{Wiki2} $\!\downarrow$
  & \textbf{PiQA} & \textbf{ArcE} & \textbf{ArcC} & \textbf{HS}
  & \textbf{WG} & \textbf{BoolQ} & \textbf{OBQA}
  & \textbf{Avg} $\!\uparrow$ \\
\midrule
\multirow{3}{*}{LLaMA-2-7B}
 & GPTQ  & W2A4 & 36.74 & 54.0 & 31.7 & 22.7 & 31.1 & 51.1 & 57.3 & 26.4 & 39.2 \\
 & GPTAQ & W2A4 & 10.91 & 61.6 & 44.7 & 25.1 & 43.3 & 55.5 & 62.1 & 30.2 & 46.1 \\
 & \cellcolor{gray!15}\textbf{\KronQ} & \cellcolor{gray!15}\textbf{W2A4} & \cellcolor{gray!15}\textbf{9.46} & \cellcolor{gray!15}\textbf{63.0} & \cellcolor{gray!15}\textbf{48.3} & \cellcolor{gray!15}\textbf{26.6} & \cellcolor{gray!15}\textbf{45.3} & \cellcolor{gray!15}\textbf{55.7} & \cellcolor{gray!15}\textbf{63.7} & \cellcolor{gray!15}\textbf{31.4} & \cellcolor{gray!15}\textbf{47.7} \\
 \midrule
\multirow{3}{*}{LLaMA-2-13B}
 & GPTQ  & W2A4 & 12.55 & 60.9 & 42.7 & 25.1 & 40.6 & 55.6 & 62.1 & 28.8 & 45.1 \\
 & GPTAQ & W2A4 &  8.41 & 66.5 & 52.0 & 29.4 & 49.4 & 58.6 & 62.5 & 35.2 & 50.5 \\
 & \cellcolor{gray!15}\textbf{\KronQ} & \cellcolor{gray!15}\textbf{W2A4} & \cellcolor{gray!15}\textbf{7.77} & \cellcolor{gray!15}\textbf{67.2} & \cellcolor{gray!15}\textbf{55.4} & \cellcolor{gray!15}\textbf{29.0} & \cellcolor{gray!15}\textbf{51.4} & \cellcolor{gray!15}\textbf{59.4} & \cellcolor{gray!15}\textbf{62.5} & \cellcolor{gray!15}\textbf{33.8} & \cellcolor{gray!15}\textbf{51.2} \\
\midrule
\multirow{3}{*}{LLaMA-3-8B}
 & GPTQ  & W2A4 & 32.79 & 56.5 & 36.7 & 21.8 & 34.6 & 52.8 & 60.0 & 27.2 & 41.4 \\
 & GPTAQ & W2A4 & 19.14 & 59.9 & 40.4 & 24.3 & 41.0 & 53.8 & 62.6 & 27.8 & 44.3 \\
 & \cellcolor{gray!15}\textbf{\KronQ} & \cellcolor{gray!15}\textbf{W2A4} & \cellcolor{gray!15}\textbf{16.47} & \cellcolor{gray!15}\textbf{60.7} & \cellcolor{gray!15}\textbf{39.7} & \cellcolor{gray!15}\textbf{25.3} & \cellcolor{gray!15}\textbf{41.4} & \cellcolor{gray!15}\textbf{53.4} & \cellcolor{gray!15}\textbf{62.6} & \cellcolor{gray!15}\textbf{30.0} & \cellcolor{gray!15}\textbf{44.7} \\

\bottomrule
\end{NiceTabular}}
 \vspace{-15pt}
\end{table}
\vspace{-4mm}
\subsection{Results on Uniform Precision}
\label{sec:exp:wonly}
\vspace{-3mm}
\noindent \textbf{Weight-only Quantization.} Table~\ref{tab:weight_only} shows WikiText-2 perplexity and zero-shot accuracy on three reasoning benchmarks (PiQA, ArcC, WG) for per-channel weight-only quantization at W4/W3/W2. The comparison baselines include OmniQuant~\citep{shao2023omniquant}, AWQ~\citep{lin2024awq}, BoA~\citep{kim2024boa}, SpinQuant~\citep{liu2024spinquant}, OSTQuant~\citep{hu2025ostquant}, QEP~\citep{arai2025quantization}, QuIP~\citep{chee2023quip}, QuIP\#~\citep{tseng2024quip},  GPTQ~\citep{frantar2022gptq}, and GPTAQ~\citep{li2025gptaq}. For QuIP\#, we report the no-E8P, no-finetuning variant to match our scalar quantization setting. \KronQ{} achieves the lowest perplexity across nearly all settings, with the most significant gains at W2 and W3 where activation-covariance-only methods degrade severely. The zero-shot accuracy improvements mirror the perplexity gains, with \KronQ{} achieving the highest or comparable accuracy across benchmarks and bit-widths. On LLaMA-3-70B, GPTQ and GPTAQ diverge or fail to produce valid quantizations, while \KronQ{} achieves 4.41 and 7.93 at W3 and W2, respectively. We attribute this to the anomalously large weight outliers unique to LLaMA-3-70B, as further analyzed in Appendix~\ref{app:llama3_70b_dist}. The remaining zero-shot results, generalization to Mistral-7B, and results on newer model families and harder benchmarks are presented in Appendix~\ref{appendix:weight_only_zs}, \ref{appendix:mistral}, and~\ref{appendix:newer_models}, respectively.

We also compare \KronQ{} with gradient-based methods, including GuidedQuant~\citep{kim2025guidedquant} and YAQA~\citep{tseng2025model}. Table~\ref{tab:guidedquant} shows that \KronQ{} matches or exceeds GuidedQuant with a plain uniform per-channel grid. Table~\ref{tab:yaqa} compares against YAQA-B under identical INT per-channel grids and no finetuning, using both their released Hessians (RedPajama) and Hessians recomputed on our calibration set (WikiText2). \KronQ{}+GPTQ denotes \KronQ{} without the drift correction. Without the drift correction, YAQA-B performs slightly better on L2-7B while \KronQ{}+GPTQ is better on L3.1-8B. The full \KronQ{}+GPTAQ achieves the lowest perplexity in every setting, indicating that a large part of the gain comes from the drift correction. An additional study that swaps only the Hessian estimator between the two methods is provided in Appendix~\ref{app:gradient_comp}.



\noindent \textbf{Group Quantization.} 
Group quantization generally achieves better performance than per-channel quantization due to its more optimized scaling factors and zero points. Table~\ref{tab:group_w2} reports results under weight-only 2-bit group quantization with 128 column blocks. \KronQ{} maintains its lead over GPTQ and GPTAQ. For example, on LLaMA-2-7B, GPTQ degrades to 274.0 PPL under group quantization while \KronQ{} achieves 7.60. The perplexity comparison with OmniQuant~\citep{shao2023omniquant} and AWQ~\citep{lin2024awq} and the full bit-width results are shown in Appendix~\ref{appendix:comparison_group} and~\ref{appendix:group}, respectively.

\noindent \textbf{Weight-and-Activation Quantization.}
For activation quantization, we apply QuaRot~\citep{ashkboos2024quarot} framework to suppress outliers. Table~\ref{tab:wa_w2} presents W2A4 results on LLaMA-2-7B/13B and LLaMA-3-8B. \KronQ{} consistently outperforms both baselines across all settings. The gains are again largest at W2A4, where \KronQ{} reduces PPL from 36.74 to 9.46 on LLaMA-2-7B. The comparison with QuaRot~\citep{ashkboos2024quarot} and SpinQuant~\citep{liu2024spinquant} and other bit-width results are presented in Appendix~\ref{appendix:comp_wa} and ~\ref{appendix:wa}, respectively.

\vspace{-3mm}
\subsection{Results on Mixed-precision}
\label{sec:exp:mp}
\begin{figure}[t]
    
    \begin{minipage}[c]{0.5\textwidth}
        \centering
        \captionof{table}{Sublayer sensitivity rankings and WikiText-2 perplexity
        on LLaMA-2-7B. Each row cumulatively upgrades one additional sublayer
        to W3 across all layers.}
        \label{tab:mp_ranking}
        \resizebox{\textwidth}{!}{%
        \begin{tabular}{llcc}
\toprule
\textbf{Score} & \textbf{Ranking} & \textbf{Avg bits} & \textbf{Wiki2}$ \!\downarrow$ \\
\midrule
baseline W2 & - & 2.00 & 8.19 \\
\midrule
\multirow{3}{*}{$\mathrm{tr}(H_G)\!\cdot\!\mathrm{tr}(H_X)$} & 1: \texttt{down\_proj}  & 2.17 & 7.21 \\
 & 2: \texttt{gate\_proj}  & 2.29 & 6.70 \\
& 3: \texttt{up\_proj}    & 2.43 & 6.38 \\
\midrule
\multirow{3}{*}{$\mathrm{tr}(H_X)$}                          & 1: \texttt{gate\_proj}  & 2.17 & 7.40 \\
& 2: \texttt{up\_proj}    & 2.29 & 6.96 \\
& 3: \texttt{down\_proj}  & 2.43 & 6.38 \\
\bottomrule
\end{tabular}}
        
    \end{minipage}
    \hfill
    \begin{minipage}[c]{0.47\textwidth}
        \centering
        \includegraphics[width=\textwidth]{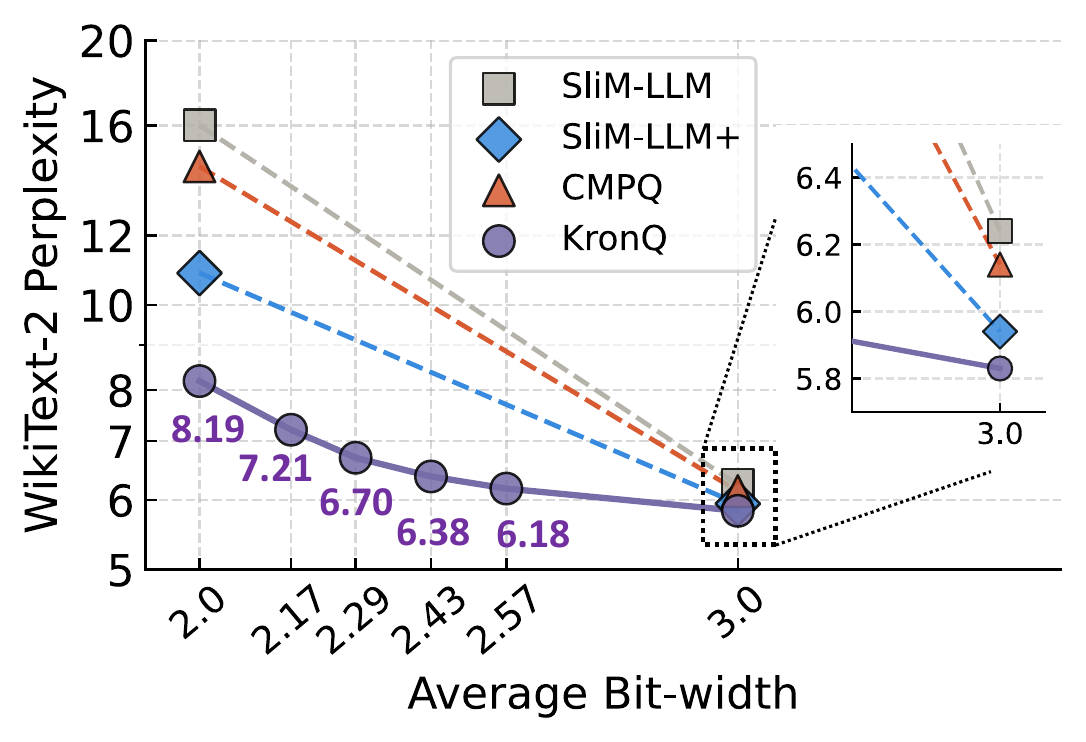}
        \vspace{-20pt}
        \caption{WikiText-2 perplexity vs.\ average bit-width on LLaMA-2-7B, comparing \KronQ{} with prior mixed precision works.}
        \label{fig:mp}
    \end{minipage}
\vspace{-6mm}
\end{figure}

\vspace{-2mm}
We evaluate the mixed-precision allocation strategy by incrementally upgrading the most sensitive sublayers from W2 to W3 according to the sensitivity score $s = \mathrm{tr}(\mathbf{H}_G) \cdot \mathrm{tr}(\mathbf{H}_X)$ in Equation~\eqref{eq:sensitivity}, applying each upgrade across all layers. Table~\ref{tab:mp_ranking} compares the sublayer rankings and WikiText-2 perplexity under the \KronQ{} joint score versus the activation-only score $\mathrm{tr}(\mathbf{H}_X)$ on LLaMA-2-7B. The activation-only score produces suboptimal allocations: its top-1 sublayer (\texttt{gate\_proj}) yields higher perplexity than the top-1 of the \KronQ{} joint score (\texttt{down\_proj}) at equal average bit-width. The \KronQ{} joint score resolves this degeneracy via $\mathbf{H}_G$, yielding strictly better PPL--bits tradeoff. Figure~\ref{fig:mp} further compares \KronQ{} mixed-precision against SliM-LLM~\citep{huang2024slim}, SliM-LLM+, and CMPQ~\citep{chen2024channel} on LLaMA-2-7B, where \KronQ{} achieves lower perplexity than W3 baselines at only ${\sim}2.6$ average bits. Other architectures show the same trend in Appendix~\ref{appn:mp}, and a comparison with inter-layer mixed-precision methods is provided in Appendix~\ref{app:mp-baselines}.


\vspace{-3mm}
\subsection{Memory and Latency}
\begin{figure*}
    \centering
    \includegraphics[width=\textwidth]{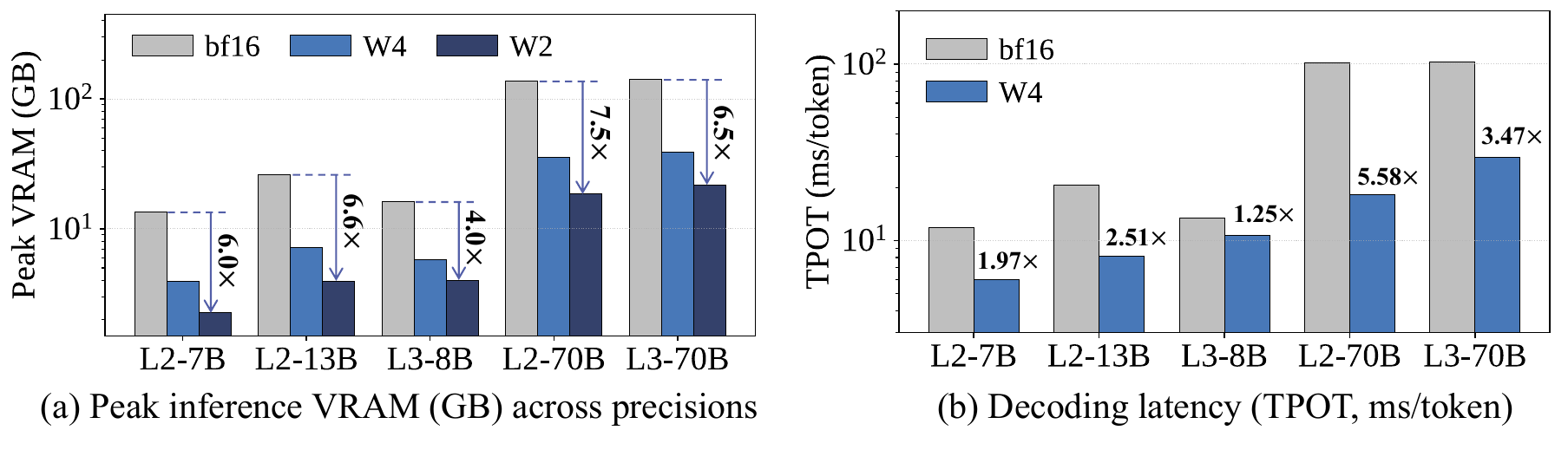}
    \vspace{-7.5mm}
    \caption{Inference efficiency of \KronQ{} quantized models. (a) Peak VRAM and (b) decoding latency (TPOT) at batch size 1, relative to the bf16 baseline. For 70B, the bf16 baseline runs on two GPUs due to OOM on single GPU, while W4 runs on a single GPU.}
\label{fig:efficiency}
\vspace{-2.1mm}
\end{figure*}

\vspace{-2mm}
Beyond calibration cost, we measure the end-to-end inference benefit of quantization in Figure~\ref{fig:efficiency}. On the memory side, \KronQ{} reduces peak inference VRAM by $3.5$--$3.9\times$ at W4 and $4.0$--$7.5\times$ at W2 over the bf16 baseline, consistently across model scales. This compression translates into deployment: a 70B model that requires two 80\,GB A100s in bf16 ($138$--$141$\,GB) fits on a single A100 at W4 ($35$--$39$\,GB). On the latency side, the reduced memory traffic yields $1.25$--$2.51\times$ faster decoding at W4, measured as time per output token (TPOT), on the 7B--13B models under a matched single-GPU configuration. We report the 70B latency, though its bf16 baseline spans two GPUs and is included only for context.

\vspace{-3mm}
\subsection{Analysis}
\vspace{-2mm}

\noindent \textbf{Ablation Study.}
We conduct three ablation studies to isolate the contribution of each component in \KronQ{}, with W2 WikiText-2 perplexity shown in Table~\ref{tab:ablation}.
First, to isolate the contribution of the base quantizer, we replace GPTAQ with a plain GPTQ base. This degrades perplexity from 8.19/6.99/11.92 to 10.18/7.95/14.52 on LLaMA-2-7B/13B and LLaMA-3-8B, confirming that drift correction and BiIP are complementary. Second, we examine the effect of diagonal rescaling ($\mathbf{S}_X$, $\mathbf{S}_G$) in BiIP. Removing the scaling while retaining the Hadamard rotations consistently degrades perplexity across all models, demonstrating that diagonal rescaling is a necessary complement to the orthogonal transforms. Third, we ablate the rotation directions in BiIP. Applying input-side incoherence ($\mathbf{H}_X$ only) already reduces perplexity, confirming the importance of suppressing input-side outliers~\citep{chee2023quip}. Output-side incoherence alone ($\mathbf{H}_G$ only) degrades severely on LLaMA-2-7B/13B as input-side outliers remain unaddressed. Combining both directions consistently achieves the lowest perplexity, demonstrating that $\mathbf{H}_G$ provides complementary output-side correction that $\mathbf{H}_X$ alone cannot capture.

\begin{table}[t]
\centering
\caption{Ablation study on WikiText-2 perplexity at W2.}
\vspace{-5pt}
\label{tab:ablation}
\resizebox{0.85\linewidth}{!}{%
\begin{NiceTabular}{l||cc|cc|ccc}
\toprule
\multirow{2}{*}{\textbf{Model}}
  & \multicolumn{2}{c}{\textbf{Base}}
  & \multicolumn{2}{c}{\textbf{Scaling} ($\mathbf{S}_X$, $\mathbf{S}_G$)}
  & \multicolumn{3}{c}{\textbf{Incoherence}} \\
  & GPTQ & GPTAQ
  & \xmark & \cmark
  & $\mathbf{H}_X$ only & $\mathbf{H}_G$ only & BiIP \\
\midrule
LLaMA-2-7B  & 10.18 & \cellcolor{gray!15}\textbf{8.19} & 8.44  & \cellcolor{gray!15}\textbf{8.19} & 8.46  & 180.43 & \cellcolor{gray!15}\textbf{8.19}  \\
LLaMA-2-13B & 7.95  & \cellcolor{gray!15}\textbf{6.99} & 7.10  & \cellcolor{gray!15}\textbf{6.99} & 7.13  & 81.18  & \cellcolor{gray!15}\textbf{6.99}  \\
LLaMA-3-8B  & 14.52 & \cellcolor{gray!15}\textbf{11.92} & 12.94 & \cellcolor{gray!15}\textbf{11.92} & 12.56 & 14.55 & \cellcolor{gray!15}\textbf{11.92} \\
\bottomrule
\end{NiceTabular}}
\vspace{-15pt}
\end{table}

\noindent \textbf{Calibration Efficiency.}

\begin{wrapfigure}{r}{0.5\textwidth}
  \centering
  \vspace{-4mm}
  \begin{minipage}{0.95\linewidth}
    \centering
    \captionof{table}{Calibration latency (s/layer) and memory needed to perform calibration (GiB) per layer. $\dagger$ represents $\mathbf{H}_G$ offload after BiIP.}
    \vspace{-1.5mm}
    \label{tab:calib_efficiency}
    \resizebox{\linewidth}{!}{
    \begin{tabular}{llcc}
      \toprule
      Model & Method & Latency & Memory \\
      \midrule
      \multirow{3}{*}{LLaMA-2-7B}
        & GPTQ  & 24.95 & 1.57 \\
        & GPTAQ & 30.11 & 1.99 \\
        & \KronQ{} & 38.25 & $3.20 \to 1.99^\dagger$ \\
      \midrule
      \multirow{3}{*}{LLaMA-3-8B}
        & GPTQ  & 26.13 & 1.97 \\
        & GPTAQ & 35.21 & 2.54 \\
        & \KronQ{} & 45.59 & $4.38 \to 2.54^\dagger$ \\
      \midrule
      \multirow{3}{*}{LLaMA-2-13B}
        & GPTQ  & 32.00 & 2.46 \\
        & GPTAQ & 39.88 & 3.11 \\
        & \KronQ{} & 50.74 & $5.02 \to 3.11^\dagger$ \\
      \bottomrule
    \end{tabular}}
  \end{minipage}
  \vspace{-4mm}
\end{wrapfigure}
\KronQ{} introduces a moderate latency overhead over GPTAQ (+8--11s/layer across models), attributable to the bidirectional Hadamard rotations in BiIP, which require dense matrix multiplications for both $\mathbf{H}_X$ and $\mathbf{H}_G$. This overhead is consistent across model sizes, reflecting that the dominant cost scales with the hidden dimension rather than the total parameter count. For memory, \KronQ{} additionally stores $\mathbf{H}_G$ as a full $d_\text{out} \times d_\text{out}$ matrix during BiIP preprocessing, causing a temporary memory overhead above GPTAQ. However, since $\mathbf{H}_G$ cancels algebraically in the column-wise quantization updates in Proposition~\ref{prop:hg_cancel}, it is released upon completion of BiIP, and the memory required for the subsequent quantization loop is identical to GPTAQ. Per-sublayer memory breakdown is provided in Appendix~\ref{app:memory}.

\vspace{-4mm}
\section{Conclusion}
\vspace{-3mm}
We presented \KronQ{}, a PTQ framework that incorporates the gradient covariance $\mathbf{H}_G$ into the quantization pipeline. \KronQ{} exploits $\mathbf{H}_G$ for bidirectional incoherence processing and inter-layer mixed-precision allocation, while inheriting the computational efficiency of GPTAQ as $\mathbf{H}_G$ cancels algebraically in the quantization updates. Experiments on LLaMA-2 and LLaMA-3 from 7B to 70B show consistent gains across W2/W3/W4, with the largest improvements at 2-bit, where activation-covariance-only methods degrade severely. The limitations of this work are presented in Appendix~\ref{limit}.


\section*{Acknowledgments}

This work was supported in part by CoCoSys, a JUMP2.0 center sponsored by DARPA and SRC, the National Science Foundation (CAREER Award, Grant \#2312366, Grant \#2318152), the DARPA Young Faculty Award, the DoE MMICC center SEA-CROGS (Award \#DE-SC0023198), and the Global Industrial Technology Cooperation Center (GITCC) program.

\bibliography{colm2026_conference}
\bibliographystyle{colm2026_conference}

\newpage
\appendix

\section{Proofs}
\label{app:proofs}

\subsection{Proof of Proposition~\ref{prop:hg_cancel} (\KronQ{} Weight Compensation)}
\label{proof_prop1}
The general OBS optimal weight update~\citep{hassibi1992second} is:
\begin{equation}
    \Delta\boldsymbol{\theta} = -\mathbf{H}^{-1}\mathbf{E}_p(\mathbf{E}_p^\top\mathbf{H}^{-1}\mathbf{E}_p)^{-1}\boldsymbol{\varepsilon}_p,
\end{equation}
where $\boldsymbol{\theta} = \mathrm{vec}(\mathbf{W})$ stacks the columns of $\mathbf{W}$, $\mathbf{E}_p = e_p \otimes \mathbf{I}$ with $e_p \in \mathbb{R}^{d_\mathrm{in}}$ and $\mathbf{I} \in \mathbb{R}^{d_\mathrm{out} \times d_\mathrm{out}}$ is the selection matrix for column $p$, and $\boldsymbol{\varepsilon}_p = \mathbf{W}_{:,p} - \widehat{\mathbf{W}}_{:,p}$ is the quantization error of column $p$. Under the K-FAC approximation $\mathbf{H} = \mathbf{H}_X \otimes \mathbf{H}_G$, we have $\mathbf{H}^{-1} = \mathbf{H}_X^{-1} \otimes \mathbf{H}_G^{-1}$.
We compute the two terms in the OBS formula. First:

\begin{align}
    \mathbf{E}_p^\top \mathbf{H}^{-1} \mathbf{E}_p
    &= (e_p^\top \otimes \mathbf{I})(\mathbf{H}_X^{-1} \otimes \mathbf{H}_G^{-1})(e_p \otimes \mathbf{I}) \nonumber \\
    &= (e_p^\top \mathbf{H}_X^{-1} e_p) \otimes \mathbf{H}_G^{-1} \nonumber \\
    &= [\mathbf{H}_X^{-1}]_{pp} \cdot \mathbf{H}_G^{-1}.
\end{align}

Second:

\begin{equation}
    \mathbf{H}^{-1}\mathbf{E}_p = (\mathbf{H}_X^{-1} \otimes \mathbf{H}_G^{-1})(e_p \otimes \mathbf{I})
    = [\mathbf{H}_X^{-1}]_{:,p} \otimes \mathbf{H}_G^{-1}.
\end{equation}

Substituting into the OBS formula and applying the mixed-product property,
$\mathbf{H}_G^{-1}$ cancels algebraically:

\begin{align}
    \Delta\boldsymbol{\theta}
    &= -\left([\mathbf{H}_X^{-1}]_{:,p} \otimes \mathbf{H}_G^{-1}\right)
       \left([\mathbf{H}_X^{-1}]_{pp} \cdot \mathbf{H}_G^{-1}\right)^{-1}
       \boldsymbol{\varepsilon}_p \nonumber \\
    &= -\frac{1}{[\mathbf{H}_X^{-1}]_{pp}}
       \left([\mathbf{H}_X^{-1}]_{:,p} \otimes \mathbf{H}_G^{-1}\mathbf{H}_G\right)
       \boldsymbol{\varepsilon}_p
    = -[\mathbf{H}_X^{-1}]_{:,p} \otimes \boldsymbol{\delta}_p,
\end{align}

where $\boldsymbol{\delta}_p = \boldsymbol{\varepsilon}_p / [\mathbf{H}_X^{-1}]_{pp}$ is the scaled quantization error of Proposition~\ref{prop:hg_cancel}.
Reshaping back to matrix form and restricting to the remaining columns gives

\begin{equation}
    \Delta\mathbf{W}_{:,p+1:}
    = -\boldsymbol{\delta}_p \cdot [\mathbf{H}_X^{-1}]_{p,p+1:}.
\end{equation}

We now extend the derivation to the full objective~\eqref{eq:kronq_full}, whose drift term contributes a linear term $\mathbf{b}^\top \Delta\boldsymbol{\theta}$ with $\mathbf{b} = 2\,\mathrm{vec}(\mathbf{H}_G \mathbf{W} \Delta\mathbf{X}\mathbf{X}^\top)$. The constrained minimizer generalizes the OBS update to

\begin{equation}
\Delta\boldsymbol{\theta} = -\tfrac{1}{2}\mathbf{H}^{-1}\mathbf{b} 
- \mathbf{H}^{-1}\mathbf{E}_p\big(\mathbf{E}_p^\top\mathbf{H}^{-1}\mathbf{E}_p\big)^{-1}
\big(\boldsymbol{\varepsilon}_p - \tfrac{1}{2}\mathbf{E}_p^\top\mathbf{H}^{-1}\mathbf{b}\big),
\label{eq:prop1_general}
\end{equation}

which reduces to the update above when $\Delta\mathbf{X} = \mathbf{0}$. Using $\mathrm{vec}(\mathbf{A}\mathbf{B}\mathbf{C}) = (\mathbf{C}^\top \otimes \mathbf{A})\,\mathrm{vec}(\mathbf{B})$,

\begin{equation}
\begin{split}
\tfrac{1}{2}\mathbf{H}^{-1}\mathbf{b} 
&= (\mathbf{H}_X^{-1} \otimes \mathbf{H}_G^{-1})\,
\mathrm{vec}\big(\mathbf{H}_G \mathbf{W} \Delta\mathbf{X}\mathbf{X}^\top\big) \\
&= \mathrm{vec}\big(\mathbf{H}_G^{-1}\mathbf{H}_G\, \mathbf{W} \Delta\mathbf{X}\mathbf{X}^\top \mathbf{H}_X^{-1}\big) 
= \mathrm{vec}\big(\mathbf{W} \Delta\mathbf{X}\mathbf{X}^\top \mathbf{H}_X^{-1}\big),
\end{split}
\label{eq:prop1_drift}
\end{equation}

so $\mathbf{H}_G$ cancels in the drift component as well, and $\mathbf{E}_p^\top\big(\tfrac{1}{2}\mathbf{H}^{-1}\mathbf{b}\big) = [\mathbf{W}\Delta\mathbf{X}\mathbf{X}^\top\mathbf{H}_X^{-1}]_{:,p}$ is likewise $\mathbf{H}_G$-free. Combined with the cancellations above, every occurrence of $\mathbf{H}_G$ cancels in~\eqref{eq:prop1_general}, so the constrained minimizer of~\eqref{eq:kronq_full} coincides with that of the unweighted ($\mathbf{H}_G = \mathbf{I}$) GPTAQ objective. Applying the sequential residual decomposition of \citet{li2025gptaq} to this system yields the correction matrix $\mathbf{P}$ and Equation~\eqref{eq:kronq_update}. $\square$

\subsection{Proof of Theorem~\ref{thm:invariance} (Rotation Invariance)}
\label{proof:theorem1}
Denote the transformed quantities after \eqref{eq:biip} as
$\mathbf{W}' = \mathbf{U}\mathbf{W}\mathbf{V}^\top$,
$\widetilde{\mathbf{H}}_G = \mathbf{U}\mathbf{H}_G\mathbf{U}^\top$,
$\widetilde{\mathbf{H}}_X = \mathbf{V}\mathbf{H}_X\mathbf{V}^\top$, and
$(\Delta\mathbf{X}\mathbf{X}^\top)' = \mathbf{V}\Delta\mathbf{X}\mathbf{X}^\top\mathbf{V}^\top$.
Note that $\Delta\mathbf{W}' = \mathbf{W}' - \widehat{\mathbf{W}}' = \mathbf{U}\Delta\mathbf{W}\mathbf{V}^\top$.
We verify invariance of each term separately.

For the first term:

\begin{align}
    \mathrm{tr}\!\left[\widetilde{\mathbf{H}}_G\,\Delta\mathbf{W}'\,
    \widetilde{\mathbf{H}}_X\,(\Delta\mathbf{W}')^\top\right]
    &= \mathrm{tr}\!\left[
        \mathbf{U}\mathbf{H}_G\mathbf{U}^\top \cdot \mathbf{U}\Delta\mathbf{W}\mathbf{V}^\top
        \cdot \mathbf{V}\mathbf{H}_X\mathbf{V}^\top \cdot \mathbf{V}\Delta\mathbf{W}^\top\mathbf{U}^\top
    \right] \nonumber \\
    &= \mathrm{tr}\!\left[
        \mathbf{U}\mathbf{H}_G\Delta\mathbf{W}\mathbf{H}_X\Delta\mathbf{W}^\top\mathbf{U}^\top
    \right] \nonumber \\
    &= \mathrm{tr}\!\left[\mathbf{H}_G\,\Delta\mathbf{W}\,\mathbf{H}_X\,\Delta\mathbf{W}^\top\right],
\end{align}

where the second step uses $\mathbf{U}^\top\mathbf{U} = \mathbf{I}$ and
$\mathbf{V}^\top\mathbf{V} = \mathbf{I}$, and the third step uses the
cyclic property of the trace.

For the second term:

\begin{align}
    \mathrm{tr}\!\left[\widetilde{\mathbf{H}}_G\,\mathbf{W}'\,
    (\Delta\mathbf{X}\mathbf{X}^\top)'\,(\Delta\mathbf{W}')^\top\right]
    &= \mathrm{tr}\!\left[
        \mathbf{U}\mathbf{H}_G\mathbf{U}^\top \cdot \mathbf{U}\mathbf{W}\mathbf{V}^\top
        \cdot \mathbf{V}\Delta\mathbf{X}\mathbf{X}^\top\mathbf{V}^\top
        \cdot \mathbf{V}\Delta\mathbf{W}^\top\mathbf{U}^\top
    \right] \nonumber \\
    &= \mathrm{tr}\!\left[
        \mathbf{U}\mathbf{H}_G\,\mathbf{W}\,\Delta\mathbf{X}\mathbf{X}^\top\,
        \Delta\mathbf{W}^\top\mathbf{U}^\top
    \right] \nonumber \\
    &= \mathrm{tr}\!\left[\mathbf{H}_G\,\mathbf{W}\,\Delta\mathbf{X}\mathbf{X}^\top\,
    \Delta\mathbf{W}^\top\right],
\end{align}

where the second step uses $\mathbf{U}^\top\mathbf{U} = \mathbf{I}$ and
$\mathbf{V}^\top\mathbf{V} = \mathbf{I}$, and the third step uses the
cyclic property of the trace.

Since both terms are individually invariant, the full \KronQ{} objective
\eqref{eq:kronq_full} is invariant under the transformation \eqref{eq:biip}.
\hfill$\square$

\subsection{Proof of Proposition~\ref{prop:biip}}
\label{proof:prop2}

\paragraph{Step 1: loss decomposition.}
By Theorem~\ref{thm:invariance} we work with the transformed quantities $\widetilde{\mathbf{H}}_X, \widetilde{\mathbf{H}}_G$ of Equations~\eqref{eq:rescale}--\eqref{eq:biip}. Let $\widetilde{\mathbf{H}}_X = \mathbf{T} \mathbf{D} \mathbf{T}^{\top}$ be the unit-triangular factorization matched to the processing order, $\mathbf{e}$ the accumulated error, and $\bm{\eta} = \mathbf{e}\,\mathbf{T}$ the per-step rounding residuals, so that $\mathbf{e}\,\widetilde{\mathbf{H}}_X\,\mathbf{e}^{\top} = \bm{\eta} \mathbf{D} \bm{\eta}^{\top}$ exactly~\citep{chee2023quip}. Hence

\begin{equation}
        L_q = \sum_{k} \mathbf{D}_{kk}\,
    \bm{\eta}_{:,k}^{\top}\,\widetilde{\mathbf{H}}_G\,\bm{\eta}_{:,k}.
    \label{eq:app-decomp}
\end{equation}

Under stochastic rounding the residuals of different rows are conditionally independent and mean-zero, since each row's feedback uses only its own past. Writing $\sigma_i^2 = \mathbb{E}[\eta_{ik}^2]$ for the per-row residual variance, constant across steps $k$,

\begin{equation}
    \mathbb{E}[L_q]
    \;=\;
    \Big(\sum_{i} [\widetilde{\mathbf{H}}_G]_{ii}\,\sigma_i^{2}\Big)
    \operatorname{tr}(\mathbf{D}),
    \label{eq:app-expected}
\end{equation}

so the off-diagonal of $\widetilde{\mathbf{H}}_G$ does not enter the expected loss.

\paragraph{Step 2: optimality of $\mathbf{S}_G$.}
The RHT mixes rows, so row energies and ranges concentrate and $\sigma_i^2 \le c_b\,\tfrac{1}{d_{\mathrm{out}}}\sum_j s_j^2 a_j$ with high probability over the sign draw, where $c_b \propto (2^b-1)^{-2}$ up to logarithmic factors~\citep{chee2023quip,tseng2024quip}. By the same concentration the bound is tight up to the deviation of row ranges from their common value, so \eqref{eq:app-product} holds as an equality to leading order. Since $\sum_i [\widetilde{\mathbf{H}}_G]_{ii} = \sum_j h_j/s_j^2$ by trace invariance, \eqref{eq:app-expected} gives

\begin{equation}
    \mathbb{E}[L_q]
    \;\le\;
    \frac{c_b}{d_{\mathrm{out}}}
    \Big(\sum_j \frac{h_j}{s_j^{2}}\Big)
    \Big(\sum_j s_j^{2} a_j\Big)
    \operatorname{tr}(\mathbf{D}).
    \label{eq:app-product}
\end{equation}

By Cauchy--Schwarz,

\begin{equation}
    \Big(\sum_j \frac{h_j}{s_j^{2}}\Big)
    \Big(\sum_j s_j^{2} a_j\Big)
    \;\ge\;
    \Big(\sum_j \sqrt{h_j\,a_j}\Big)^{\!2},
    \label{eq:app-cs}
\end{equation}

with equality iff $s_j \propto (h_j/a_j)^{1/4}$, the $\mathbf{S}_G$ of Equation~\eqref{eq:rescale} up to a global scale that leaves the product invariant. This yields Equation~\eqref{eq:biip-bound}. The gain over $s \equiv 1$ is $\rho = \big(\sum_j\sqrt{h_j a_j}\big)^2 / \big(\sum_j h_j \sum_j a_j\big) \le 1$, and $\rho \approx k/d_{\mathrm{out}}$ when the curvature concentrates on $k$ channels of comparable row energy. The random signs also decorrelate the residuals from $\widetilde{\mathbf{H}}_G$, extending \eqref{eq:app-expected} to nearest rounding in expectation over the sign draw. \hfill$\square$

\paragraph{Relation to \citet{tseng2025model}.}
\citet{tseng2025model} bound the proxy loss of their two-sided solver by $\operatorname{tr}(\mathbf{D}_X)\operatorname{tr}(\mathbf{D}_G)\sigma^{2}$, where the output-side factor enters the solver as feedback. Ours covers the one-sided solver \KronQ{} runs, where $\mathbf{H}_G$ acts only as a diagonal preconditioner.

\paragraph{Numerical verification.}
On Llama-2-7B layer-0 matrices with INT4 per-channel grids, \eqref{eq:app-decomp} holds to machine precision, $\mathbb{E}[L_q]$ matches \eqref{eq:app-expected} within $0.6\%$, and the exponent sweep $s_i=(h_i/a_i)^{\alpha}$ is minimized at $\alpha=1/4$. For nearest rounding the diagonal-only prediction holds within $2.3\%$ averaged over RHT sign draws, though with larger per-draw deviation. This analysis covers per-channel grids and the GPTQ-mode objective, without GPTAQ's drift term.

\begin{table}[h]
\centering
\small
\caption{Expected proxy loss under the one-sided solver, up to the grid constant $c_b/d_{\mathrm{out}}$. All rows include the BiIP rotations. The first two omit the rescaling ($s\equiv 1$).}
\label{tab:optimality}
\begin{tabular}{lc}
\toprule
\textbf{Method} & \textbf{Expected proxy loss factor} \\
\midrule
Nearest rounding & $\big(\sum_i h_i\big)\big(\sum_i a_i\big)\operatorname{tr}(\widetilde{\mathbf{H}}_X)$ \\
GPTQ & $\big(\sum_i h_i\big)\big(\sum_i a_i\big)\operatorname{tr}(\mathbf{D})$ \\
\KronQ{} & $\big(\sum_i \sqrt{h_i a_i}\big)^{2}\operatorname{tr}(\mathbf{D})$ \\
\bottomrule
\end{tabular}
\end{table}

\section{\KronQ{} Algorithm}
\label{appendix:alg}

\begin{algorithm}[H]
\caption{Bidirectional Incoherence Pre-Processing (BiIP)}
\label{algo1}
\small
\begin{algorithmic}[1]
\Require FP weight $\mathbf{W} \in \mathbb{R}^{d_\mathrm{out} \times d_\mathrm{in}}$,
         degraded input $\mathbf{X} \in \mathbb{R}^{d_\mathrm{in} \times n}$,
         FP input $\tilde{\mathbf{X}} \in \mathbb{R}^{d_\mathrm{in} \times n}$,
         $\mathbf{H}_G \in \mathbb{R}^{d_\mathrm{out} \times d_\mathrm{out}}$ (precomputed)
\State $\mathbf{H}_X \leftarrow \mathbf{X}\mathbf{X}^\top$;\quad
       $\Delta\mathbf{X}\mathbf{X}^\top \leftarrow (\tilde{\mathbf{X}} - \mathbf{X})\mathbf{X}^\top$
       \Comment{online statistics}
\State \textbf{sample} random Hadamard $\mathbf{U} \in \mathbb{R}^{d_\mathrm{out} \times d_\mathrm{out}}$,\;
       $\mathbf{V} \in \mathbb{R}^{d_\mathrm{in} \times d_\mathrm{in}}$
\State $\mathbf{S}_X \leftarrow \bigl(\mathrm{diag}(\mathbf{H}_X) / \mathrm{diag}(\mathbf{W}^\top \mathbf{W})\bigr)^{1/4}$
       \Comment{input-side rescaling}
\State $\mathbf{S}_G \leftarrow \bigl(\mathrm{diag}(\mathbf{H}_G) / \mathrm{diag}(\mathbf{W}\mathbf{W}^\top)\bigr)^{1/4}$
       \Comment{output-side rescaling}
\State $\mathbf{W} \leftarrow \mathbf{S}_G\,\mathbf{W}\,\mathbf{S}_X$
\State $\mathbf{H}_X \leftarrow \mathbf{S}_X^{-1}\mathbf{H}_X\mathbf{S}_X^{-1}$;\quad
       $\Delta\mathbf{X}\mathbf{X}^\top \leftarrow \mathbf{S}_X^{-1}\Delta\mathbf{X}\mathbf{X}^\top\mathbf{S}_X^{-1}$;\quad
        $\mathbf{H}_G \leftarrow \mathbf{S}_G^{-1}\mathbf{H}_G\mathbf{S}_G^{-1}$
\State $\mathbf{W} \leftarrow \mathbf{U}\mathbf{W}\mathbf{V}^\top$;\quad
       $\mathbf{H}_X \leftarrow \mathbf{V}\mathbf{H}_X\mathbf{V}^\top$;\quad
       $\Delta\mathbf{X}\mathbf{X}^\top \leftarrow \mathbf{V}\,\Delta\mathbf{X}\mathbf{X}^\top\mathbf{V}^\top$
       \Comment{incoherence}
\State $\mathbf{H}_G \leftarrow \mathbf{U}\mathbf{H}_G\mathbf{U}^\top$;\quad \textbf{release} $\mathbf{H}_G$
       \Comment{$\mathbf{H}_G$ not needed after BiIP (Prop.~1)}
\State \textbf{return} $\mathbf{W},\;\mathbf{H}_X,\;\Delta\mathbf{X}\mathbf{X}^\top,\;\mathbf{S}_X,\;\mathbf{S}_G,\;\mathbf{U},\;\mathbf{V}$
\end{algorithmic}
\end{algorithm}

\begin{algorithm}[H]
\caption{\KronQ{} -- Quantization Loop}
\label{algo2}

\small
\begin{algorithmic}[1]
\Require $\mathbf{W} \in \mathbb{R}^{m \times n},\,
          \mathbf{H}_X \in \mathbb{R}^{n \times n},\,
          \Delta\mathbf{X}\mathbf{X}^\top \in \mathbb{R}^{n \times n},\,
          \mathbf{S}_X,\, \mathbf{S}_G,\, \mathbf{U},\, \mathbf{V}$
         (from Alg.~\ref{algo1}),
         block size $B$, damping $\lambda$, scaling $\alpha$
\State $\mathbf{H}_X \leftarrow \mathbf{H}_X + \lambda \cdot \mathrm{mean}(\mathrm{diag}(\mathbf{H}_X))\,\mathbf{I}$
       \Comment{damping}
\State $\mathbf{L} \leftarrow \textit{Inverse\_Cholesky}(\mathbf{H}_X)$
\State $\mathbf{P} \leftarrow \alpha\bigl((\Delta\mathbf{X}\mathbf{X}^\top \cdot \mathbf{L}) \odot \mathbf{M}_U\bigr)\mathbf{L}^\top$
       \Comment{GPTAQ correction; $\mathbf{M}_U$: upper-triangular mask}
\State $\mathbf{Q} \leftarrow \mathbf{0}_{m \times n}$;\quad
       $\mathbf{E} \leftarrow \mathbf{0}_{m \times B}$
\For{$i = 0,\, B,\, 2B,\, \ldots$}
    \For{$j = i,\; i{+}1,\; \ldots,\; i{+}B{-}1$}
        \State $\mathbf{Q}_{:,j} \leftarrow \mathrm{quant}(\mathbf{W}_{:,j})$
        \State $\mathbf{E}_{:,\,j-i} \leftarrow (\mathbf{W}_{:,j} - \mathbf{Q}_{:,j})\;/\;\mathbf{L}_{jj}$
        \State $\mathbf{W}_{:,\,j:(i+B)} \leftarrow \mathbf{W}_{:,\,j:(i+B)}
                - \mathbf{E}_{:,\,j-i}\,\mathbf{L}_{j,\,j:(i+B)}^\top
                + \mathbf{W}_{:,j}\,\mathbf{P}_{j,\,j:(i+B)}$
    \EndFor
    \State $\mathbf{W}_{:,\,(i+B):} \leftarrow \mathbf{W}_{:,\,(i+B):}
            - \mathbf{E}\cdot\mathbf{L}_{i:(i+B),\,(i+B):}^\top
            + \mathbf{W}_{:,\,i:(i+B)}\,\mathbf{P}_{i:(i+B),\,(i+B):}$
\EndFor
\State $\mathbf{Q} \leftarrow \mathbf{U}^\top \mathbf{Q}\,\mathbf{V}$;\quad
       $\mathbf{Q} \leftarrow \mathbf{S}_G^{-1}\,\mathbf{Q}\,\mathbf{S}_X^{-1}$
       \Comment{revert incoherence \& rescaling}
\State \textbf{return} $\mathbf{Q}$
\end{algorithmic}
\end{algorithm}

\section{Additional Experiments}

\subsection{Zero-shot Accuracy on Weight-only Quantization}
\label{appendix:weight_only_zs}

\begingroup
\fontsize{7.3pt}{8.6pt}\selectfont
\setlength{\tabcolsep}{6pt}
\begin{longtable}{l||l|c|cccccccr}
\caption{Zero-shot accuracy on reasoning benchmarks (PiQA, ArcE, ArcC, HS, WG, BoolQ, OBQA) for weight-only quantization (WxA16) at W4/W3/W2. NaN indicates diverged quantization.}
\label{tab:weight_only_zs} \\

\hline
\textbf{Model} & \textbf{Method} & \textbf{Bits}
  & \textbf{PiQA} & \textbf{ArcE} & \textbf{ArcC} & \textbf{HS}
  & \textbf{WG} & \textbf{BoolQ} & \textbf{OBQA}
  & \textbf{Avg} $\!\uparrow$ \\
\hline
\endfirsthead

\multicolumn{11}{c}{{\tablename\ \thetable{} -- continued}} \\
\hline
\textbf{Model} & \textbf{Method} & \textbf{Bits}
  & \textbf{PiQA} & \textbf{ArcE} & \textbf{ArcC} & \textbf{HS}
  & \textbf{WG} & \textbf{BoolQ} & \textbf{OBQA}
  & \textbf{Avg} $\!\uparrow$ \\
\hline
\endhead

\multicolumn{11}{r}{{Continued on next page}} \\
\endfoot

\hline
\endlastfoot

\multirow{13}{*}{LLaMA-2-7B}
 & GPTQ  & 4 & 77.2 & 69.2 & 41.2 & 70.5 & 68.7 & 67.3 & 40.8 & 62.1 \\
 & GPTAQ & 4 & 76.8 & 70.6 & 41.6 & 70.3 & 68.4 & 68.1 & 42.6 & 62.6 \\
 & \cellcolor{gray!15}\textbf{\KronQ} & \cellcolor{gray!15}\textbf{4} & \cellcolor{gray!15}\textbf{78.7} & \cellcolor{gray!15}\textbf{72.4} & \cellcolor{gray!15}\textbf{44.9} & \cellcolor{gray!15}\textbf{75.2} & \cellcolor{gray!15}\textbf{68.4} & \cellcolor{gray!15}\textbf{77.4} & \cellcolor{gray!15}\textbf{44.4} & \cellcolor{gray!15}\textbf{65.9} \\
\cline{2-11}
 & OmniQuant & 3 & 74.0 & 62.6 & 36.3 & 67.6 & 63.6 & 65.7 & 39.8 & 58.5 \\
 & GPTQ      & 3 & 76.0 & 59.4 & 36.5 & 66.8 & 65.7 & 65.9 & 40.0 & 58.6 \\
 & BoA       & 3 & 78.4 & 69.3 & 40.3 & 72.0 & 67.6 & 74.2 & 41.2 & 63.3 \\
 & GPTAQ     & 3 & 75.4 & 62.0 & 37.2 & 65.9 & 65.3 & 66.1 & 38.4 & 58.6 \\
 & \cellcolor{gray!15}\textbf{\KronQ} & \cellcolor{gray!15}\textbf{3} & \cellcolor{gray!15}\textbf{76.9} & \cellcolor{gray!15}\textbf{70.9} & \cellcolor{gray!15}\textbf{43.3} & \cellcolor{gray!15}\textbf{72.5} & \cellcolor{gray!15}\textbf{67.2} & \cellcolor{gray!15}\textbf{72.5} & \cellcolor{gray!15}\textbf{41.4} & \cellcolor{gray!15}\textbf{63.5} \\
\cline{2-11}
 & OmniQuant & 2 & 57.2 & 35.3 & 26.1 & 31.7 & 51.5 & 51.1 & 30.6 & 40.5 \\
 & GPTQ      & 2 & 63.0 & 45.1 & 28.4 & 42.8 & 53.9 & 52.3 & 28.4 & 44.8 \\
 & BoA       & 2 & 64.9 & 47.7 & 28.3 & 45.5 & 54.5 & 64.7 & 29.6 & 47.9 \\
 & GPTAQ     & 2 & NaN & NaN & NaN & NaN & NaN & NaN & NaN & NaN \\
 & \cellcolor{gray!15}\textbf{\KronQ} & \cellcolor{gray!15}\textbf{2} & \cellcolor{gray!15}\textbf{68.0} & \cellcolor{gray!15}\textbf{54.3} & \cellcolor{gray!15}\textbf{31.3} & \cellcolor{gray!15}\textbf{51.2} & \cellcolor{gray!15}\textbf{60.8} & \cellcolor{gray!15}\textbf{64.3} & \cellcolor{gray!15}\textbf{34.8} & \cellcolor{gray!15}\textbf{52.0} \\
\hline
\multirow{13}{*}{LLaMA-2-13B}
 & GPTQ  & 4 & 77.3 & 73.7 & 45.6 & 67.6 & 71.1 & 71.0 & 43.2 & 64.2 \\
 & GPTAQ & 4 & 77.9 & 74.4 & 46.5 & 73.1 & 71.3 & 75.5 & 44.8 & 66.2 \\
 & \cellcolor{gray!15}\textbf{\KronQ} & \cellcolor{gray!15}\textbf{4} & \cellcolor{gray!15}\textbf{80.2} & \cellcolor{gray!15}\textbf{74.7} & \cellcolor{gray!15}\textbf{48.0} & \cellcolor{gray!15}\textbf{79.1} & \cellcolor{gray!15}\textbf{71.8} & \cellcolor{gray!15}\textbf{82.1} & \cellcolor{gray!15}\textbf{44.8} & \cellcolor{gray!15}\textbf{68.7} \\
\cline{2-11}
 & OmniQuant & 3 & 77.7 & 69.0 & 42.7 & 72.8 & 65.9 & 69.0 & 38.8 & 62.3 \\
 & GPTQ      & 3 & 72.5 & 62.8 & 36.8 & 54.4 & 65.5 & 66.5 & 36.6 & 56.4 \\
 & BoA       & 3 & 77.4 & 69.3 & 43.5 & 74.7 & 62.5 & 78.9 & 35.2 & 63.1 \\
 & GPTAQ     & 3 & 75.1 & 65.5 & 39.0 & 62.6 & 64.2 & 67.3 & 38.2 & 58.8 \\
 & \cellcolor{gray!15}\textbf{\KronQ} & \cellcolor{gray!15}\textbf{3} & \cellcolor{gray!15}\textbf{79.3} & \cellcolor{gray!15}\textbf{73.5} & \cellcolor{gray!15}\textbf{46.5} & \cellcolor{gray!15}\textbf{77.0} & \cellcolor{gray!15}\textbf{71.2} & \cellcolor{gray!15}\textbf{78.9} & \cellcolor{gray!15}\textbf{45.0} & \cellcolor{gray!15}\textbf{67.3} \\
\cline{2-11}
 & OmniQuant & 2 & 62.9 & 44.4 & 28.4 & 49.7 & 52.6 & 62.2 & 33.8 & 47.7 \\
 & GPTQ      & 2 & 57.9 & 36.2 & 23.4 & 37.3 & 53.2 & 62.1 & 28.2 & 42.6 \\
 & BoA       & 2 & 63.3 & 46.3 & 29.8 & 49.3 & 52.6 & 62.2 & 27.6 & 47.3 \\
 & GPTAQ     & 2 & 54.8 & 33.1 & 21.4 & 31.7 & 52.0 & 58.0 & 29.4 & 40.1 \\
 & \cellcolor{gray!15}\textbf{\KronQ} & \cellcolor{gray!15}\textbf{2} & \cellcolor{gray!15}\textbf{72.4} & \cellcolor{gray!15}\textbf{62.5} & \cellcolor{gray!15}\textbf{36.4} & \cellcolor{gray!15}\textbf{59.4} & \cellcolor{gray!15}\textbf{65.1} & \cellcolor{gray!15}\textbf{69.4} & \cellcolor{gray!15}\textbf{37.0} & \cellcolor{gray!15}\textbf{57.5} \\
\hline
\multirow{11}{*}{LLaMA-2-70B}
 & GPTQ  & 4 & 82.3 & 81.0 & 57.6 & 82.7 & 77.4 & 82.3 & 48.0 & 73.0 \\
 & GPTAQ & 4 & 82.8 & 81.5 & 57.3 & 82.8 & 77.0 & 82.4 & 47.8 & 73.1 \\
 & \cellcolor{gray!15}\textbf{\KronQ} & \cellcolor{gray!15}\textbf{4} & \cellcolor{gray!15}\textbf{82.3} & \cellcolor{gray!15}\textbf{80.3} & \cellcolor{gray!15}\textbf{57.9} & \cellcolor{gray!15}\textbf{83.8} & \cellcolor{gray!15}\textbf{79.9} & \cellcolor{gray!15}\textbf{85.4} & \cellcolor{gray!15}\textbf{49.2} & \cellcolor{gray!15}\textbf{74.1} \\
\cline{2-11}
 & OmniQuant & 3 & 80.7 & 75.6 & 45.8 & 78.1 & 73.6 & 66.5 & 42.6 & 66.1 \\
 & GPTQ      & 3 & 80.4 & 78.9 & 55.8 & 80.2 & 76.2 & 79.1 & 46.0 & 70.9 \\
 & GPTAQ     & 3 & 81.7 & 77.5 & 54.4 & 79.8 & 75.2 & 81.6 & 45.8 & 70.9 \\
 & \cellcolor{gray!15}\textbf{\KronQ} & \cellcolor{gray!15}\textbf{3} & \cellcolor{gray!15}\textbf{82.3} & \cellcolor{gray!15}\textbf{81.9} & \cellcolor{gray!15}\textbf{58.3} & \cellcolor{gray!15}\textbf{82.5} & \cellcolor{gray!15}\textbf{78.5} & \cellcolor{gray!15}\textbf{83.0} & \cellcolor{gray!15}\textbf{48.8} & \cellcolor{gray!15}\textbf{73.6} \\
\cline{2-11}
 & OmniQuant & 2 & 69.9 & 55.5 & 31.5 & 55.4 & 53.6 & 64.9 & 33.8 & 52.1 \\
 & GPTQ      & 2 & 72.1 & 64.7 & 37.6 & 58.0 & 66.9 & 66.6 & 40.2 & 58.0 \\
 & GPTAQ     & 2 & 72.2 & 64.4 & 36.8 & 58.4 & 64.6 & 69.9 & 34.6 & 57.3 \\
 & \cellcolor{gray!15}\textbf{\KronQ} & \cellcolor{gray!15}\textbf{2} & \cellcolor{gray!15}\textbf{77.4} & \cellcolor{gray!15}\textbf{75.9} & \cellcolor{gray!15}\textbf{46.9} & \cellcolor{gray!15}\textbf{72.0} & \cellcolor{gray!15}\textbf{75.0} & \cellcolor{gray!15}\textbf{79.7} & \cellcolor{gray!15}\textbf{43.2} & \cellcolor{gray!15}\textbf{67.2} \\
\hline
\multirow{11}{*}{LLaMA-3-8B}
 & GPTQ  & 4 & 79.4 & 75.7 & 49.7 & 75.6 & 73.5 & 77.1 & 44.8 & 68.0 \\
 & GPTAQ & 4 & 80.0 & 75.7 & 49.7 & 77.2 & 72.5 & 76.5 & 44.4 & 68.0 \\
 & \cellcolor{gray!15}\textbf{\KronQ} & \cellcolor{gray!15}\textbf{4} & \cellcolor{gray!15}\textbf{79.4} & \cellcolor{gray!15}\textbf{78.2} & \cellcolor{gray!15}\textbf{51.5} & \cellcolor{gray!15}\textbf{78.2} & \cellcolor{gray!15}\textbf{73.2} & \cellcolor{gray!15}\textbf{82.2} & \cellcolor{gray!15}\textbf{44.2} & \cellcolor{gray!15}\textbf{69.6} \\
\cline{2-11}
 & GPTQ  & 3 & 74.5 & 62.6 & 39.7 & 70.0 & 67.8 & 73.3 & 38.6 & 60.9 \\
 & BoA   & 3 & 77.3 & 72.8 & 45.1 & 72.7 & 71.4 & 78.7 & 42.6 & 65.8 \\
 & GPTAQ & 3 & 73.5 & 60.1 & 40.4 & 70.8 & 69.9 & 75.7 & 41.0 & 61.6 \\
 & \cellcolor{gray!15}\textbf{\KronQ} & \cellcolor{gray!15}\textbf{3} & \cellcolor{gray!15}\textbf{77.5} & \cellcolor{gray!15}\textbf{74.5} & \cellcolor{gray!15}\textbf{50.2} & \cellcolor{gray!15}\textbf{74.9} & \cellcolor{gray!15}\textbf{71.7} & \cellcolor{gray!15}\textbf{81.1} & \cellcolor{gray!15}\textbf{41.2} & \cellcolor{gray!15}\textbf{67.3} \\
\cline{2-11}
 & GPTQ  & 2 & 54.2 & 34.0 & 22.3 & 37.9 & 49.5 & 45.9 & 28.0 & 38.8 \\
 & BoA   & 2 & 59.9 & 44.9 & 26.6 & 43.3 & 55.8 & 60.5 & 29.4 & 45.8 \\
 & GPTAQ & 2 & 55.0 & 33.1 & 23.6 & 38.1 & 53.4 & 46.8 & 28.0 & 39.7 \\
 & \cellcolor{gray!15}\textbf{\KronQ} & \cellcolor{gray!15}\textbf{2} & \cellcolor{gray!15}\textbf{67.0} & \cellcolor{gray!15}\textbf{50.7} & \cellcolor{gray!15}\textbf{30.6} & \cellcolor{gray!15}\textbf{49.0} & \cellcolor{gray!15}\textbf{61.1} & \cellcolor{gray!15}\textbf{66.1} & \cellcolor{gray!15}\textbf{33.0} & \cellcolor{gray!15}\textbf{51.1} \\
\hline
\multirow{9}{*}{LLaMA-3-70B}
 & GPTQ  & 4 & 50.2 & 26.3 & 24.6 & 59.8 & 56.4 & 66.9 & 33.2 & 45.3 \\
 & GPTAQ & 4 & 56.9 & 39.6 & 25.6 & 30.1 & 54.4 & 48.1 & 29.0 & 40.5 \\
 & \cellcolor{gray!15}\textbf{\KronQ} & \cellcolor{gray!15}\textbf{4} & \cellcolor{gray!15}\textbf{84.2} & \cellcolor{gray!15}\textbf{81.8} & \cellcolor{gray!15}\textbf{61.7} & \cellcolor{gray!15}\textbf{84.8} & \cellcolor{gray!15}\textbf{79.1} & \cellcolor{gray!15}\textbf{86.8} & \cellcolor{gray!15}\textbf{48.0} & \cellcolor{gray!15}\textbf{75.2} \\
\cline{2-11}
 & GPTQ  & 3 & 52.0 & 25.3 & 26.1 & 26.4 & 49.0 & 41.2 & 28.4 & 35.5 \\
 & GPTAQ & 3 & 50.0 & 25.9 & 27.5 & 26.2 & 49.7 & 37.9 & 30.0 & 35.3 \\
 & \cellcolor{gray!15}\textbf{\KronQ} & \cellcolor{gray!15}\textbf{3} & \cellcolor{gray!15}\textbf{83.5} & \cellcolor{gray!15}\textbf{80.4} & \cellcolor{gray!15}\textbf{58.8} & \cellcolor{gray!15}\textbf{83.1} & \cellcolor{gray!15}\textbf{79.5} & \cellcolor{gray!15}\textbf{85.4} & \cellcolor{gray!15}\textbf{47.2} & \cellcolor{gray!15}\textbf{74.0} \\
\cline{2-11}
 & GPTQ  & 2 & 51.7 & 25.6 & 26.4 & 26.4 & 49.0 & 37.8 & 27.4 & 34.9 \\
 & GPTAQ & 2 & NaN & NaN & NaN & NaN & NaN & NaN & NaN & NaN \\
 & \cellcolor{gray!15}\textbf{\KronQ} & \cellcolor{gray!15}\textbf{2} & \cellcolor{gray!15}\textbf{77.6} & \cellcolor{gray!15}\textbf{73.2} & \cellcolor{gray!15}\textbf{44.5} & \cellcolor{gray!15}\textbf{66.0} & \cellcolor{gray!15}\textbf{72.6} & \cellcolor{gray!15}\textbf{79.1} & \cellcolor{gray!15}\textbf{39.6} & \cellcolor{gray!15}\textbf{64.7} \\
\hline
\end{longtable}
\vspace{-5mm}
\endgroup

\subsection{Comparison on Mistral-7B}
\label{appendix:mistral}
Table~\ref{tab:mistral} extends the weight-only quantization evaluation to Mistral-7B~\citep{jiang2024mistral}, demonstrating that \KronQ{} generalizes beyond the LLaMA family. 


\begin{table}[H]
\centering
\caption{Weight-only quantization (WxA16) WikiText-2 perplexity $\downarrow$ on Mistral-7B.}
\label{tab:mistral}
\footnotesize
\setlength{\tabcolsep}{4pt}
\resizebox{0.50\linewidth}{!}{%
\begin{NiceTabular}{lcccccc}[code-before = \rectanglecolor{gray!15}{2-7}{5-7}]
\toprule
 & AWQ & QEP & QuIP & GPTQ & GPTAQ & \textbf{\KronQ} \\
\midrule
W4 & 5.72 & 5.48 & 11.11 & 8.96    & 6.95  & \textbf{5.32} \\
W3 & 7.90 & 5.84 & 7.11  & 532.46  & 12.89 & \textbf{5.54} \\
W2 & NaN  & 9.59 & 26.63 & 5900.69 & 27.07 & \textbf{7.78} \\
\bottomrule
\end{NiceTabular}%
}
\end{table}

\subsection{Generalization to Newer Models and Harder Benchmarks}
\label{appendix:newer_models}
To verify that \KronQ{}'s gains are not specific to the LLaMA family or to perplexity, we evaluate recent model families: Gemma-3-12B~\citep{gemma3}, DeepSeek-R1-Distill-Llama-8B~\citep{guo2025deepseek}, and Phi-4-mini-instruct~\citep{abouelenin2025phi}. Table~\ref{tab:newer_models} reports weight-only WikiText-2 perplexity at W4 and W2 against GPTQ~\citep{frantar2022gptq} and GPTAQ~\citep{li2025gptaq}. \KronQ{} wins all eight settings, and the margin widens sharply at W2. Beyond perplexity, Table~\ref{tab:harder_bench} evaluates four harder benchmarks, GPQA-Diamond~\citep{rein2023gpqa}, MMLU~\citep{hendrycks2020measuring}, AIME-2024~\citep{maxwelljia2024aime}, and LiveCodeBench~\citep{jain2025livecodebench}, on DeepSeek-R1-Distill-Llama-8B and Gemma-3-12B-IT at W4. \KronQ{} outperforms both baselines on all four benchmarks, with the largest gains on generative reasoning tasks such as LiveCodeBench, where it nearly doubles the GPTAQ scores on both models.

\paragraph{Evaluation Details.}
All models are quantized to W4 (per-channel, asymmetric, weight-only) and calibrated on 128 WikiText-2 sequences with no fine-tuning, using identical settings across methods. MMLU, GPQA-Diamond, and AIME-2024 are evaluated through the lm-evaluation-harness~\citep{eval-harness} in the zero-shot setting, while LiveCodeBench uses its official runner with vLLM. We report log-likelihood accuracy on MMLU (57 subjects) and GPQA-Diamond (198 questions), exact-match accuracy on AIME-2024 (30 problems), and pass@1 on LiveCodeBench (\texttt{release\_v5}, code-generation scenario, 32{,}768 maximum new tokens). For DeepSeek-R1-Distill-Llama-8B we follow the decoding protocol of \citet{liu2025quantization}, using temperature 0.6, top-p 0.95, and 32{,}768 maximum new tokens with stochastic sampling, reporting AIME-2024 as avg@8. Gemma-3-12B-IT is decoded greedily with a 2{,}048-token budget, reporting AIME-2024 as pass@1.
\begin{table}[h]
\centering
\begin{minipage}[t]{0.47\linewidth}
\centering
\caption{WikiText-2 perplexity ($\downarrow$) on newer 2025 model families (WxA16, W4/W2).}
\label{tab:newer_models}
\resizebox{\linewidth}{!}{%
\begin{NiceTabular}{l||c|c|c|c}[code-before = \rectanglecolor{gray!15}{2-5}{9-5}]
\toprule
\textbf{Model} & \textbf{Bits} & \textbf{GPTQ} & \textbf{GPTAQ} & \textbf{\KronQ} \\
\midrule
\multirow{2}{*}{Gemma-3-12B-PT}
 & W4 & 39.77 & 34.84 & \textbf{27.97} \\
 & W2 & 1.3e4 & 269.0 & \textbf{60.10} \\
\midrule
\multirow{2}{*}{Gemma-3-12B-IT}
 & W4 & 60.37 & 28.82 & \textbf{25.60} \\
 & W2 & 7.1e3 & 94.08 & \textbf{54.39} \\
\midrule
\multirow{2}{*}{DeepSeek-R1-Distill-Llama-8B}
 & W4 & 16.56 & 15.32 & \textbf{13.58} \\
 & W2 & 73.53 & 110.34 & \textbf{27.72} \\
\midrule
\multirow{2}{*}{Phi-4-mini-instruct}
 & W4 & 13.81 & 11.80 & \textbf{10.17} \\
 & W2 & 111.42 & 31.07 & \textbf{19.26} \\
\bottomrule
\end{NiceTabular}}
\end{minipage}
\hfill
\begin{minipage}[t]{0.50\linewidth}
\centering
\caption{Accuracy ($\uparrow$, \%) on harder benchmarks (W4A16).}
\label{tab:harder_bench}
\vspace{4mm}
\resizebox{\linewidth}{!}{%
\begin{NiceTabular}{l|l||c|c|c}[code-before = \rectanglecolor{gray!15}{2-5}{10-5}]
\toprule
\textbf{Model} & \textbf{Benchmark} & \textbf{GPTQ} & \textbf{GPTAQ} & \textbf{\KronQ} \\
\midrule
\multirow{4}{*}{\shortstack[l]{DeepSeek-R1-\\Distill-Llama-8B}}
 & LiveCodeBench & 19.9 & 16.3 & \textbf{37.3} \\
 & AIME-2024     & 10.0 & 13.3 & \textbf{16.7} \\
 & MMLU          & 44.15 & 40.51 & \textbf{53.17} \\
 & GPQA-Diamond  & 28.80 & 27.27 & \textbf{29.29} \\
\midrule
\multirow{4}{*}{Gemma-3-12B-IT}
 & LiveCodeBench & 7.8 & 9.6 & \textbf{22.3} \\
 & AIME-2024     & 10.0 & 10.0 & \textbf{23.3} \\
 & MMLU          & 64.82 & 65.21 & \textbf{70.02} \\
 & GPQA-Diamond  & 32.83 & 34.34 & \textbf{38.38} \\
\bottomrule
\end{NiceTabular}}
\end{minipage}
\end{table}

\subsection{Comparison of \KronQ{} and YAQA Hessians}
\label{app:gradient_comp}
To separate the effect of the Hessian estimator from that of the rounding algorithm, we fix \KronQ{}'s pipeline (\KronQ{}+GPTQ) and swap only the Hessian pair. The drift correction is disabled in this study, since its statistics are collected from the activations observed during sequential quantization and cannot be paired consistently with an external $\mathbf{H}_X$. ``\KronQ{}'' is our estimator computed on the WikiText-2 calibration set. The other two rows use YAQA's Sketch-B Hessians, either recomputed on the same WikiText-2 tokens or as released (RedPajama). Table~\ref{tab:estimator_swap} shows that neither estimator dominates. Sketch-B is better on L2-7B and the K-FAC estimate is better on L3.1-8B, with the largest gaps at W2, while at W3 and W4 the differences are small.

\begin{table}[h]
\centering
\caption{Hessian estimator swap under a fixed pipeline (\KronQ{}+GPTQ,
WikiText-2 PPL, L2-7B ctx 2048, L3.1-8B ctx 8192).}
\label{tab:estimator_swap}
\scalebox{0.85}{%
\begin{tabular}{lcccccc}
\toprule
& \multicolumn{3}{c}{L2-7B} & \multicolumn{3}{c}{L3.1-8B} \\
\cmidrule(lr){2-4}\cmidrule(lr){5-7}
Estimator & W2 & W3 & W4 & W2 & W3 & W4 \\
\midrule
\KronQ{} (WikiText2)  & 10.18 & 5.91 & 5.58 & \textbf{14.32} & 7.36 & 6.72 \\
YAQA-B (WikiText2)       & \textbf{9.11} & \textbf{5.84} & \textbf{5.57} & 16.90 & \textbf{7.35} & \textbf{6.70} \\
YAQA-B (RedPajama)       & 10.77 & 5.92 & 5.59 & 21.97 & 7.36 & 6.72 \\
\bottomrule
\end{tabular}%
}
\end{table}

\subsection{Comparison with Previous Works on Group Quantization}
\label{appendix:comparison_group}

Table~\ref{tab:wiki2_group} reports WikiText-2 perplexity under group quantization ($g{=}128$), where OmniQuant~\citep{shao2023omniquant} and AWQ~\citep{lin2024awq} results are taken from their papers. 


\begin{table}[H]
\centering
\caption{Weight-only group quantization (WxA16, g=128) WikiText-2 perplexity.}
\label{tab:wiki2_group}
\resizebox{0.7\linewidth}{!}{
\begin{tabular}{l|ccc|ccc|ccc}
\toprule
& \multicolumn{3}{c|}{\textbf{LLaMA-2-7B}} 
& \multicolumn{3}{c|}{\textbf{LLaMA-2-13B}} 
& \multicolumn{3}{c}{\textbf{LLaMA-2-70B}} \\
\textbf{Method} & \textbf{W4} & \textbf{W3} & \textbf{W2} 
               & \textbf{W4} & \textbf{W3} & \textbf{W2} 
               & \textbf{W4} & \textbf{W3} & \textbf{W2} \\
\midrule
RTN       & 5.72 & 6.66 & 4.2e3  & 4.98 & 5.51 & 122.08 & 3.46 & 3.97 & 27.27 \\
GPTQ      & 5.57 & 6.02 & 274.00 & 4.95 & 5.22 &   8.40 & 3.39 & 3.64 &  5.22 \\
AWQ       & 5.60 & 6.24 & NaN    & 4.97 & 5.32 &    NaN & 3.41 & 3.74 &    -- \\
OmniQuant & 5.58 & 6.03 &  11.06 & 4.95 & 5.28 &   8.26 & 3.40 & 3.78 &  6.55 \\
GPTAQ     & 5.56 & 5.88 &  23.19 & 4.94 & 5.16 &   7.17 & 3.39 & 3.64 &  5.28 \\
\cellcolor{gray!15}\textbf{\KronQ{}} 
& \cellcolor{gray!15}\textbf{5.54} 
& \cellcolor{gray!15}\textbf{5.77} 
& \cellcolor{gray!15}\textbf{7.60} 
& \cellcolor{gray!15}\textbf{4.93} 
& \cellcolor{gray!15}\textbf{5.13} 
& \cellcolor{gray!15}\textbf{6.51} 
& \cellcolor{gray!15}\textbf{3.38} 
& \cellcolor{gray!15}\textbf{3.60} 
& \cellcolor{gray!15}\textbf{4.85} \\
\bottomrule
\end{tabular}}
\end{table}

\subsection{Additional Bit-width Results on Group Quantization}
\label{appendix:group}
Table~\ref{tab:group_full} provides full W2/W3/W4 results under group quantization ($g{=}128$). \KronQ{} maintains its lead across all bit-widths and model sizes, with particularly large margins at W2 where per-channel methods struggle.
\begingroup
\fontsize{7.3pt}{8.6pt}\selectfont
\setlength{\tabcolsep}{6pt}
\begin{longtable}{l||l|c|c|cccccccr}
\caption{Group quantization (WxA16, $g{=}128$), full results (W2/W3/W4).}
\label{tab:group_full} \\

\hline
\textbf{Model} & \textbf{Method} & \textbf{Bits}
  & \textbf{Wiki2}$\!\downarrow$
  & \textbf{PiQA} & \textbf{ArcE} & \textbf{ArcC} & \textbf{HS}
  & \textbf{WG} & \textbf{BoolQ} & \textbf{OBQA}
  & \textbf{Avg}$\!\uparrow$ \\
\hline
\endfirsthead

\multicolumn{12}{c}{{\tablename\ \thetable{} -- continued}} \\
\hline
\textbf{Model} & \textbf{Method} & \textbf{Bits}
  & \textbf{Wiki2}$\!\downarrow$
  & \textbf{PiQA} & \textbf{ArcE} & \textbf{ArcC} & \textbf{HS}
  & \textbf{WG} & \textbf{BoolQ} & \textbf{OBQA}
  & \textbf{Avg}$\!\uparrow$ \\
\hline
\endhead

\multicolumn{12}{r}{{Continued on next page}} \\
\endfoot

\hline
\endlastfoot

\multirow{6}{*}{LLaMA-2-7B}
 & GPTQ  & 4 & 5.57 & 78.6 & 72.6 & 44.7 & 75.2 & 69.0 & 77.8 & 44.0 & 66.0 \\
 & GPTAQ & 4 & 5.56 & 79.1 & 73.8 & 44.5 & 75.2 & 69.0 & 76.4 & 43.6 & 65.9 \\
 & \cellcolor{gray!15}\textbf{\KronQ} & \cellcolor{gray!15}\textbf{4} & \cellcolor{gray!15}\textbf{5.54} & \cellcolor{gray!15}\textbf{78.0} & \cellcolor{gray!15}\textbf{73.1} & \cellcolor{gray!15}\textbf{44.6} & \cellcolor{gray!15}\textbf{75.4} & \cellcolor{gray!15}\textbf{68.9} & \cellcolor{gray!15}\textbf{77.9} & \cellcolor{gray!15}\textbf{43.6} & \cellcolor{gray!15}\textbf{65.9} \\
\cline{2-12}
 & GPTQ  & 3 & 6.02 & 76.7 & 68.6 & 41.3 & 71.8 & 67.9 & 72.9 & 41.2 & 62.9 \\
 & GPTAQ & 3 & 5.88 & 77.3 & 69.7 & 41.2 & 72.4 & 68.7 & 66.9 & 40.6 & 62.4 \\
 & \cellcolor{gray!15}\textbf{\KronQ} & \cellcolor{gray!15}\textbf{3} & \cellcolor{gray!15}\textbf{5.77} & \cellcolor{gray!15}\textbf{77.8} & \cellcolor{gray!15}\textbf{73.2} & \cellcolor{gray!15}\textbf{44.5} & \cellcolor{gray!15}\textbf{73.0} & \cellcolor{gray!15}\textbf{67.1} & \cellcolor{gray!15}\textbf{72.9} & \cellcolor{gray!15}\textbf{43.4} & \cellcolor{gray!15}\textbf{64.6} \\
\hline
\multirow{6}{*}{LLaMA-3-8B}
 & GPTQ  & 4 & 6.41 & 80.3 & 78.2 & 52.0 & 78.3 & 73.6 & 80.9 & 44.4 & 69.7 \\
 & GPTAQ & 4 & 6.38 & 80.0 & 76.8 & 51.9 & 78.3 & 73.8 & 78.6 & 44.0 & 69.1 \\
 & \cellcolor{gray!15}\textbf{\KronQ} & \cellcolor{gray!15}\textbf{4} & \cellcolor{gray!15}\textbf{6.34} & \cellcolor{gray!15}\textbf{80.4} & \cellcolor{gray!15}\textbf{78.9} & \cellcolor{gray!15}\textbf{53.2} & \cellcolor{gray!15}\textbf{78.4} & \cellcolor{gray!15}\textbf{73.3} & \cellcolor{gray!15}\textbf{80.9} & \cellcolor{gray!15}\textbf{45.8} & \cellcolor{gray!15}\textbf{70.1} \\
\cline{2-12}
 & GPTQ  & 3 & 7.42 & 77.9 & 74.2 & 48.8 & 61.2 & 71.5 & 79.2 & 41.6 & 64.9 \\
 & GPTAQ & 3 & 7.51 & 77.0 & 73.3 & 46.4 & 72.0 & 72.0 & 76.8 & 42.0 & 65.6 \\
 & \cellcolor{gray!15}\textbf{\KronQ} & \cellcolor{gray!15}\textbf{3} & \cellcolor{gray!15}\textbf{6.96} & \cellcolor{gray!15}\textbf{78.7} & \cellcolor{gray!15}\textbf{74.2} & \cellcolor{gray!15}\textbf{49.1} & \cellcolor{gray!15}\textbf{74.9} & \cellcolor{gray!15}\textbf{71.7} & \cellcolor{gray!15}\textbf{81.7} & \cellcolor{gray!15}\textbf{42.4} & \cellcolor{gray!15}\textbf{67.5} \\
\hline
\multirow{6}{*}{LLaMA-2-13B}
 & GPTQ  & 4 & 4.95 & 80.2 & 76.4 & 48.2 & 78.5 & 72.1 & 79.4 & 45.0 & 68.5 \\
 & GPTAQ & 4 & 4.94 & 79.7 & 76.5 & 48.0 & 78.7 & 72.7 & 79.0 & 44.0 & 68.4 \\
 & \cellcolor{gray!15}\textbf{\KronQ} & \cellcolor{gray!15}\textbf{4} & \cellcolor{gray!15}\textbf{4.93} & \cellcolor{gray!15}\textbf{80.3} & \cellcolor{gray!15}\textbf{75.7} & \cellcolor{gray!15}\textbf{48.6} & \cellcolor{gray!15}\textbf{79.0} & \cellcolor{gray!15}\textbf{72.1} & \cellcolor{gray!15}\textbf{80.8} & \cellcolor{gray!15}\textbf{45.0} & \cellcolor{gray!15}\textbf{68.8} \\
\cline{2-12}
 & GPTQ  & 3 & 5.22 & 79.2 & 74.2 & 46.6 & 76.0 & 70.9 & 78.2 & 43.4 & 66.9 \\
 & GPTAQ & 3 & 5.16 & 79.1 & 75.3 & 46.4 & 76.1 & 71.1 & 77.7 & 44.4 & 67.2 \\
 & \cellcolor{gray!15}\textbf{\KronQ} & \cellcolor{gray!15}\textbf{3} & \cellcolor{gray!15}\textbf{5.13} & \cellcolor{gray!15}\textbf{79.2} & \cellcolor{gray!15}\textbf{76.3} & \cellcolor{gray!15}\textbf{48.7} & \cellcolor{gray!15}\textbf{77.5} & \cellcolor{gray!15}\textbf{72.1} & \cellcolor{gray!15}\textbf{81.6} & \cellcolor{gray!15}\textbf{44.6} & \cellcolor{gray!15}\textbf{68.6} \\
\hline
\multirow{9}{*}{LLaMA-2-70B}
 & GPTQ  & 4 & 3.39 & 82.8 & 80.9 & 57.2 & 83.5 & 78.1 & 82.7 & 48.0 & 73.3 \\
 & GPTAQ & 4 & 3.39 & 82.7 & 80.6 & 57.5 & 83.4 & 77.9 & 83.6 & 48.6 & 73.5 \\
 & \cellcolor{gray!15}\textbf{\KronQ} & \cellcolor{gray!15}\textbf{4} & \cellcolor{gray!15}\textbf{3.38} & \cellcolor{gray!15}\textbf{82.7} & \cellcolor{gray!15}\textbf{79.2} & \cellcolor{gray!15}\textbf{56.2} & \cellcolor{gray!15}\textbf{84.1} & \cellcolor{gray!15}\textbf{79.8} & \cellcolor{gray!15}\textbf{85.0} & \cellcolor{gray!15}\textbf{48.4} & \cellcolor{gray!15}\textbf{73.6} \\
\cline{2-12}
 & GPTQ  & 3 & 3.64 & 82.3 & 80.0 & 55.6 & 82.5 & 75.5 & 79.1 & 44.6 & 71.4 \\
 & GPTAQ & 3 & 3.64 & 82.4 & 80.8 & 55.5 & 82.2 & 76.9 & 83.2 & 47.6 & 72.7 \\
 & \cellcolor{gray!15}\textbf{\KronQ} & \cellcolor{gray!15}\textbf{3} & \cellcolor{gray!15}\textbf{3.60} & \cellcolor{gray!15}\textbf{82.2} & \cellcolor{gray!15}\textbf{80.7} & \cellcolor{gray!15}\textbf{57.1} & \cellcolor{gray!15}\textbf{83.0} & \cellcolor{gray!15}\textbf{79.0} & \cellcolor{gray!15}\textbf{83.8} & \cellcolor{gray!15}\textbf{47.8} & \cellcolor{gray!15}\textbf{73.4} \\
\cline{2-12}
 & GPTQ  & 2 & 5.22 & 77.7 & 73.3 & 47.1 & 68.4 & 71.0 & 76.0 & 39.0 & 64.6 \\
 & GPTAQ & 2 & 5.28 & 78.4 & 75.3 & 46.3 & 69.1 & 73.0 & 75.5 & 41.6 & 65.6 \\
 & \cellcolor{gray!15}\textbf{\KronQ} & \cellcolor{gray!15}\textbf{2} & \cellcolor{gray!15}\textbf{4.85} & \cellcolor{gray!15}\textbf{79.2} & \cellcolor{gray!15}\textbf{75.8} & \cellcolor{gray!15}\textbf{49.8} & \cellcolor{gray!15}\textbf{74.5} & \cellcolor{gray!15}\textbf{76.6} & \cellcolor{gray!15}\textbf{78.1} & \cellcolor{gray!15}\textbf{44.2} & \cellcolor{gray!15}\textbf{68.3} \\
\hline
\multirow{9}{*}{LLaMA-3-70B}
 & GPTQ  & 4 & 3.23 & 83.5 & 84.7 & 62.5 & 84.3 & 79.6 & 85.8 & 47.6 & 75.4 \\
 & GPTAQ & 4 & 3.17 & 83.2 & 82.9 & 61.3 & 84.3 & 80.7 & 85.4 & 48.4 & 75.2 \\
 & \cellcolor{gray!15}\textbf{\KronQ} & \cellcolor{gray!15}\textbf{4} & \cellcolor{gray!15}\textbf{3.15} & \cellcolor{gray!15}\textbf{83.7} & \cellcolor{gray!15}\textbf{80.7} & \cellcolor{gray!15}\textbf{60.7} & \cellcolor{gray!15}\textbf{85.0} & \cellcolor{gray!15}\textbf{80.7} & \cellcolor{gray!15}\textbf{87.0} & \cellcolor{gray!15}\textbf{47.6} & \cellcolor{gray!15}\textbf{75.1} \\
\cline{2-12}
 & GPTQ  & 3 & 4.49 & 82.1 & 80.0 & 58.4 & 82.6 & 78.8 & 85.1 & 44.6 & 73.1 \\
 & GPTAQ & 3 & 6.56 & 81.7 & 78.8 & 54.6 & 77.5 & 74.4 & 77.0 & 43.4 & 69.6 \\
 & \cellcolor{gray!15}\textbf{\KronQ} & \cellcolor{gray!15}\textbf{3} & \cellcolor{gray!15}\textbf{4.20} & \cellcolor{gray!15}\textbf{82.2} & \cellcolor{gray!15}\textbf{79.5} & \cellcolor{gray!15}\textbf{59.0} & \cellcolor{gray!15}\textbf{83.1} & \cellcolor{gray!15}\textbf{78.9} & \cellcolor{gray!15}\textbf{84.9} & \cellcolor{gray!15}\textbf{47.6} & \cellcolor{gray!15}\textbf{73.6} \\
\cline{2-12}
 & GPTQ  & 2 & 16.60 & 74.2 & 68.3 & 42.6 & 40.2 & 69.3 & 74.7 & 38.4 & 58.2 \\
 & GPTAQ & 2 & 20.62 & 56.4 & 37.9 & 24.7 & 37.3 & 50.7 & 53.6 & 28.2 & 41.3 \\
 & \cellcolor{gray!15}\textbf{\KronQ} & \cellcolor{gray!15}\textbf{2} & \cellcolor{gray!15}\textbf{7.65} & \cellcolor{gray!15}\textbf{79.0} & \cellcolor{gray!15}\textbf{76.4} & \cellcolor{gray!15}\textbf{49.7} & \cellcolor{gray!15}\textbf{72.2} & \cellcolor{gray!15}\textbf{73.8} & \cellcolor{gray!15}\textbf{80.7} & \cellcolor{gray!15}\textbf{43.4} & \cellcolor{gray!15}\textbf{67.9} \\
\hline
\end{longtable}
\endgroup

\subsection{Comparison with Previous Works on Weight-and-activation Quantization}
\label{appendix:comp_wa}
Table~\ref{tab:wa_quant} reports WikiText-2 perplexity for weight-and-activation quantization. All methods apply the QuaRot~\citep{ashkboos2024quarot} rotation framework for activation quantization. QuaRot and SpinQuant~\citep{liu2024spinquant} results are taken from their respective papers. 

\begin{table}[H]
\centering
\caption{Weight-and-activation quantization (WxA4) WikiText-2 perplexity.}
\label{tab:wa_quant}
\resizebox{0.6\linewidth}{!}{
\begin{tabular}{ll|ccc}
\toprule
\textbf{Bits} & \textbf{Method} 
& \textbf{LLaMA-2-7B} & \textbf{LLaMA-2-13B} & \textbf{LLaMA-3-8B} \\
\midrule
\multirow{5}{*}{W4A4}
& QuaRot    &  6.10 &  5.40  &  8.16 \\
& SpinQuant &  5.96 &  5.24 &  7.39 \\
& GPTQ      &  6.04 &  5.29  &  7.78 \\
& GPTAQ     &  5.87 &  5.17 &  7.39 \\
& \cellcolor{gray!15}\textbf{\KronQ} & \cellcolor{gray!15}\textbf{5.82} & \cellcolor{gray!15}\textbf{5.15} & \cellcolor{gray!15}\textbf{7.34} \\
\midrule
\multirow{4}{*}{W2A4}
& QuaRot    & 32.6  & 15.5   & NaN   \\
& GPTQ      & 36.74 & 12.55  & 32.79 \\
& GPTAQ     & 10.91 &  8.41   & 19.14 \\
& \cellcolor{gray!15}\textbf{\KronQ} & \cellcolor{gray!15}\textbf{9.46} & \cellcolor{gray!15}\textbf{7.77} & \cellcolor{gray!15}\textbf{16.47} \\
\bottomrule
\end{tabular}}
\end{table}

\subsection{3 and 4-bit Results on Weight-and-activation Quantization}
\label{appendix:wa}
Table~\ref{tab:wa_full} extends the weight-and-activation evaluation to W3A4/W4A4. \KronQ{} achieves the lowest perplexity across all settings, with gains increasing at lower bit-widths.
\begin{table}[H]
\caption{Weight-and-activation quantization (WxA4), W3A4/W4A4 results.}
\label{tab:wa_full}
\centering
\resizebox{0.9\linewidth}{!}{%
\begin{NiceTabular}{l||l|c|c|ccccccccr}
\toprule
\textbf{Model} & \textbf{Method} & \textbf{Bits}
  & \textbf{Wiki2}$\!\downarrow$
  & \textbf{PiQA} & \textbf{ArcE} & \textbf{ArcC} & \textbf{HS}
  & \textbf{WG} & \textbf{BoolQ} & \textbf{OBQA}
  & \textbf{Avg}$\!\uparrow$ \\
\midrule
LLaMA-2-7B
 & GPTQ  & W4A4 &  6.04 & 77.3 & 71.0 & 42.2 & 72.9 & 66.1 & 73.5 & 39.2 & 63.2 \\
 & GPTAQ & W4A4 &  5.87 & 77.2 & 71.6 & 43.2 & 73.6 & 67.2 & 74.5 & 41.6 & 64.1 \\
 & \cellcolor{gray!15}\textbf{\KronQ} & \cellcolor{gray!15}\textbf{W4A4} & \cellcolor{gray!15}\textbf{5.82} & \cellcolor{gray!15}\textbf{77.2} & \cellcolor{gray!15}\textbf{71.2} & \cellcolor{gray!15}\textbf{42.3} & \cellcolor{gray!15}\textbf{73.8} & \cellcolor{gray!15}\textbf{67.6} & \cellcolor{gray!15}\textbf{73.8} & \cellcolor{gray!15}\textbf{40.8} & \cellcolor{gray!15}\textbf{63.8} \\
\cmidrule{2-14}
 & GPTQ  & W3A4 &  6.79 & 75.0 & 67.8 & 40.2 & 68.4 & 63.1 & 71.7 & 36.6 & 60.4 \\
 & GPTAQ & W3A4 &  6.25 & 75.3 & 68.3 & 40.4 & 70.5 & 65.8 & 71.9 & 40.2 & 61.8 \\
 & \cellcolor{gray!15}\textbf{\KronQ} & \cellcolor{gray!15}\textbf{W3A4} & \cellcolor{gray!15}\textbf{6.18} & \cellcolor{gray!15}\textbf{76.8} & \cellcolor{gray!15}\textbf{67.5} & \cellcolor{gray!15}\textbf{40.2} & \cellcolor{gray!15}\textbf{70.2} & \cellcolor{gray!15}\textbf{66.6} & \cellcolor{gray!15}\textbf{73.0} & \cellcolor{gray!15}\textbf{41.4} & \cellcolor{gray!15}\textbf{62.2} \\
\midrule
LLaMA-2-13B
 & GPTQ  & W4A4 &  5.29 & 78.9 & 74.7 & 47.2 & 76.5 & 69.6 & 79.0 & 44.4 & 67.2 \\
 & GPTAQ & W4A4 &  5.17 & 79.3 & 75.2 & 48.6 & 77.0 & 68.8 & 78.7 & 43.4 & 67.3 \\
 & \cellcolor{gray!15}\textbf{\KronQ} & \cellcolor{gray!15}\textbf{W4A4} & \cellcolor{gray!15}\textbf{5.15} & \cellcolor{gray!15}\textbf{79.2} & \cellcolor{gray!15}\textbf{75.6} & \cellcolor{gray!15}\textbf{48.4} & \cellcolor{gray!15}\textbf{77.7} & \cellcolor{gray!15}\textbf{70.5} & \cellcolor{gray!15}\textbf{80.3} & \cellcolor{gray!15}\textbf{44.2} & \cellcolor{gray!15}\textbf{68.0} \\
\cmidrule{2-14}
 & GPTQ  & W3A4 &  5.77 & 77.4 & 72.5 & 44.8 & 73.3 & 69.1 & 71.4 & 42.0 & 64.4 \\
 & GPTAQ & W3A4 &  5.48 & 77.7 & 73.1 & 44.6 & 74.0 & 69.9 & 74.7 & 42.8 & 65.3 \\
 & \cellcolor{gray!15}\textbf{\KronQ} & \cellcolor{gray!15}\textbf{W3A4} & \cellcolor{gray!15}\textbf{5.40} & \cellcolor{gray!15}\textbf{79.1} & \cellcolor{gray!15}\textbf{75.0} & \cellcolor{gray!15}\textbf{45.8} & \cellcolor{gray!15}\textbf{75.0} & \cellcolor{gray!15}\textbf{70.2} & \cellcolor{gray!15}\textbf{77.2} & \cellcolor{gray!15}\textbf{43.8} & \cellcolor{gray!15}\textbf{66.6} \\
\midrule
LLaMA-3-8B
 & GPTQ  & W4A4 &  7.78 & 77.0 & 72.1 & 45.7 & 74.2 & 67.9 & 76.4 & 42.6 & 65.1 \\
 & GPTAQ & W4A4 &  7.39 & 77.4 & 72.0 & 47.5 & 75.6 & 69.0 & 77.5 & 42.0 & 65.9 \\
 & \cellcolor{gray!15}\textbf{\KronQ} & \cellcolor{gray!15}\textbf{W4A4} & \cellcolor{gray!15}\textbf{7.34} & \cellcolor{gray!15}\textbf{77.4} & \cellcolor{gray!15}\textbf{75.3} & \cellcolor{gray!15}\textbf{48.7} & \cellcolor{gray!15}\textbf{75.1} & \cellcolor{gray!15}\textbf{70.1} & \cellcolor{gray!15}\textbf{75.8} & \cellcolor{gray!15}\textbf{42.4} & \cellcolor{gray!15}\textbf{66.4} \\
\cmidrule{2-14}
 & GPTQ  & W3A4 &  9.22 & 73.3 & 64.2 & 39.3 & 68.8 & 64.3 & 72.8 & 39.2 & 60.3 \\
 & GPTAQ & W3A4 &  8.31 & 74.7 & 68.6 & 42.1 & 70.3 & 65.9 & 73.4 & 40.8 & 62.3 \\
 & \cellcolor{gray!15}\textbf{\KronQ} & \cellcolor{gray!15}\textbf{W3A4} & \cellcolor{gray!15}\textbf{8.25} & \cellcolor{gray!15}\textbf{74.6} & \cellcolor{gray!15}\textbf{69.5} & \cellcolor{gray!15}\textbf{42.7} & \cellcolor{gray!15}\textbf{70.0} & \cellcolor{gray!15}\textbf{67.6} & \cellcolor{gray!15}\textbf{76.0} & \cellcolor{gray!15}\textbf{38.8} & \cellcolor{gray!15}\textbf{62.7} \\
\bottomrule
\end{NiceTabular}}
\end{table}

\subsection{Mixed Precision Quantization on LLaMA}
\label{appn:mp}
\begin{figure}[h]
\centering
\begin{minipage}[t]{0.50\textwidth}
\vspace{0pt}
\centering
\captionof{table}{Sublayer sensitivity rankings and WikiText-2 perplexity on LLaMA-3-8B and LLaMA-2-13B. Each row cumulatively upgrades one additional sublayer to W3 across all layers.}
\label{tab:mixed_precision_appendix}
\resizebox{\linewidth}{!}{
\begin{tabular}{llcc}
\toprule
\textbf{Score} & \textbf{Ranking} & \textbf{Avg bits} & \textbf{Wiki2}$\!\downarrow$ \\
\midrule
\multicolumn{4}{l}{\textit{LLaMA-3-8B}} \\
\midrule
baseline W2 & - & 2.00 & 11.92 \\
\midrule
\multirow{3}{*}{$\mathrm{tr}(H_G)\!\cdot\!\mathrm{tr}(H_X)$} & 1: \texttt{down\_proj} & 2.17 & 9.88 \\
 & 2: \texttt{gate\_proj} & 2.29 & 8.67 \\
 & 3: \texttt{up\_proj}   & 2.43 & 7.85 \\
\midrule
\multirow{3}{*}{$\mathrm{tr}(H_X)$} & 1: \texttt{gate\_proj} & 2.17 & 10.24 \\
 & 2: \texttt{up\_proj}   & 2.29 & 9.16 \\
 & 3: \texttt{down\_proj} & 2.43 & 7.85 \\
\midrule
\multicolumn{4}{l}{\textit{LLaMA-2-13B}} \\
\midrule
baseline W2 & - & 2.00 & 6.99 \\
\midrule
\multirow{3}{*}{$\mathrm{tr}(H_G)\!\cdot\!\mathrm{tr}(H_X)$} & 1: \texttt{down\_proj} & 2.17 & 6.29 \\
 & 2: \texttt{gate\_proj} & 2.29 & 5.86 \\
 & 3: \texttt{up\_proj}   & 2.43 & 5.56 \\
\midrule
\multirow{3}{*}{$\mathrm{tr}(H_X)$} & 1: \texttt{gate\_proj} & 2.17 & 6.39 \\
 & 2: \texttt{up\_proj}   & 2.29 & 6.00 \\
 & 3: \texttt{down\_proj} & 2.43 & 5.56 \\
\bottomrule
\end{tabular}}
\end{minipage}
\hfill
\begin{minipage}[t]{0.48\textwidth}
\vspace{0pt}
\centering
\includegraphics[width=\linewidth]{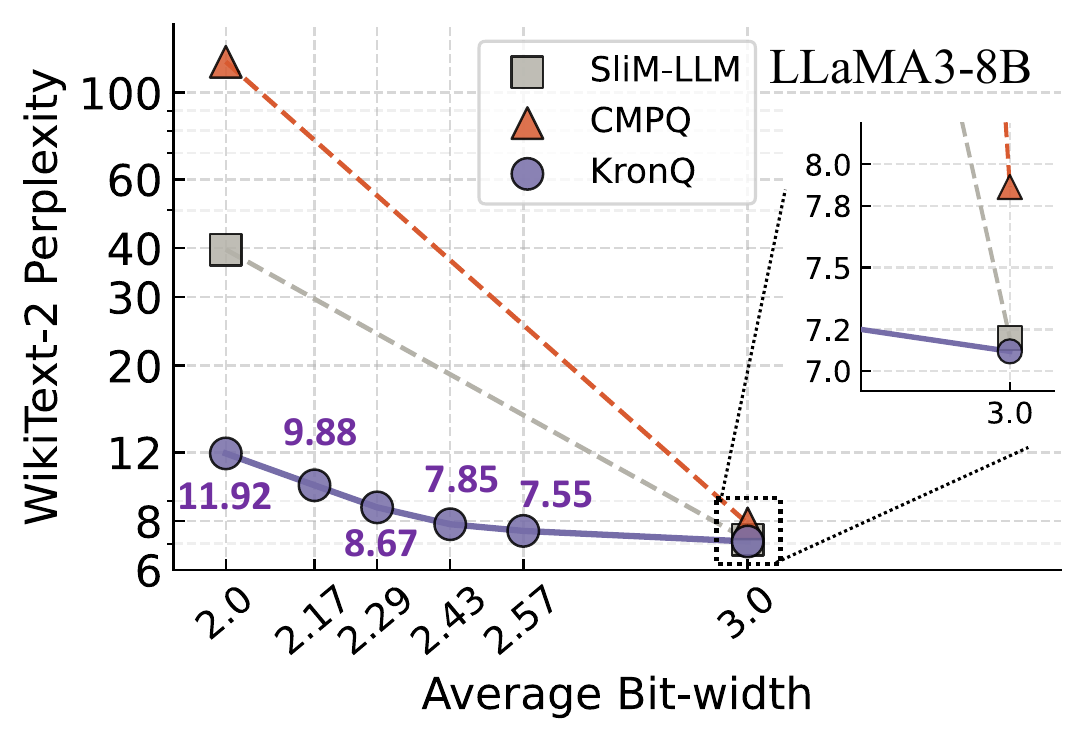}
\label{fig:mp_llama3_8b}
\vspace{-3mm}
\includegraphics[width=\linewidth]{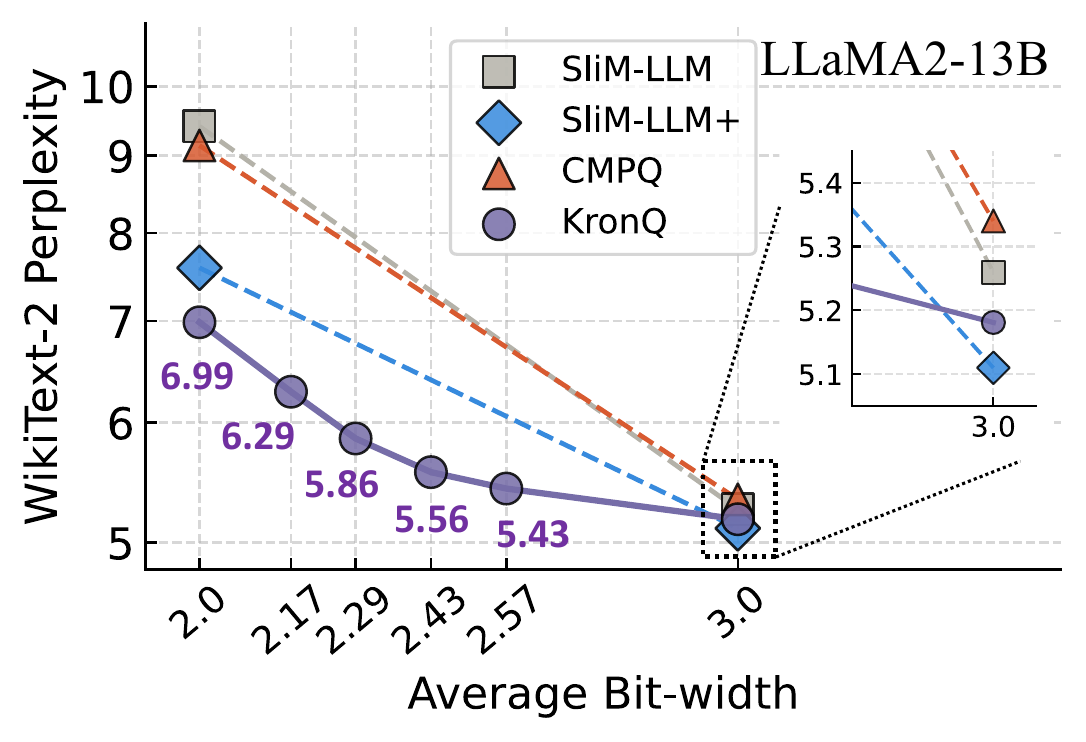}
\label{fig:mp_llama2_13b}
\vspace{-5mm}
\end{minipage}
\end{figure}

\vspace{-1mm}
Table~\ref{tab:mixed_precision_appendix} and right figures extend the mixed-precision analysis of Section~\ref{sec:exp:mp} to LLaMA-3-8B and LLaMA-2-13B. The \KronQ{} joint score yields strictly better PPL--bits tradeoffs than the activation-only score on both architectures. We also compare \KronQ{} with SliM-LLM~\citep{huang2024slim} and CMPQ~\citep{chen2024channel}, consistent with the LLaMA-2-7B results in the main paper.

\vspace{-5mm}
\subsection{Comparison with Inter-Layer Mixed-Precision Quantization}
\label{app:mp-baselines}
\vspace{-1mm}
Since \KronQ{} allocates bit-widths across sublayers, we further compare against the inter-layer methods AMQ~\citep{lee2025amq}, Q-Palette~\citep{lee2026q}, and HIGGS~\citep{malinovskii2025higgs}. All comparisons are conducted under a matched data-aware setting: every method uses the same real-text calibration set (128 WikiText-2 sequences of length 2048) that \KronQ{} uses for $\mathbf{H_G}$, so the comparison isolates the allocation methodology from calibration-data access.

For the allocation comparisons, we hold the base quantizer fixed and vary only the allocation. Table~\ref{tab:amq} holds GPTQ fixed (LLaMA-2-7B, context 2048), where \KronQ{}'s allocation beats AMQ's data-aware NSGA-II search at a matched 3.1-bit budget. Table~\ref{tab:qpalette} holds Q-Palette's quantizer fixed (LLaMA-3.1-8B, context 8192) and compares against Q-Palette's data-aware actual-loss allocation. \KronQ{}'s allocation achieves lower perplexity at every budget. Table~\ref{tab:higgs} instead isolates the quantizer: against the data-aware GPTQ+HIGGS configuration (LLaMA-2-7B, context 4096), \KronQ{}'s scalar grid leads at every bit-width despite HIGGS using a vector quantizer.
Crucially, \KronQ{}'s allocation is a closed-form analytic score that needs only the single $\mathbf{H_G}$ backward pass already computed for BiIP, with no dedicated search or per-bit precompute. It is therefore free within the \KronQ{} pipeline, versus roughly 5 GPU-hours for AMQ's NSGA-II search and 79 GPU-hours for Q-Palette's data-aware actual-loss term. \KronQ{} thus matches the data-aware allocation quality at roughly one twentieth of the cost.

\vspace{-1mm}
\begin{table}[h]
\centering
\begin{minipage}[t]{0.30\linewidth}
\centering
\caption{\KronQ{} vs. AMQ allocation, GPTQ fixed (L2-7B, ctx 2048).}
\label{tab:amq}
\vspace{2pt}\footnotesize
\begin{NiceTabular}{lcc}[code-before = \rectanglecolor{gray!15}{3-1}{3-3}]
\toprule
Alloc. & Bits & PPL \\
\midrule
AMQ & 3.1 & 6.68 \\
\textbf{\KronQ{}} & 3.1 & \textbf{6.53} \\
\bottomrule
\end{NiceTabular}
\end{minipage}\hfill
\begin{minipage}[t]{0.34\linewidth}
\centering
\caption{\KronQ{} vs. Q-Palette allocation, Q-Palette quantizer fixed (L3.1-8B, ctx 8192).}
\label{tab:qpalette}
\vspace{2pt}\footnotesize
\begin{NiceTabular}{lcc}[code-before = \rectanglecolor{gray!15}{2-3}{4-3}]
\toprule
Bits & Q-Palette & \KronQ{} \\
\midrule
3.19 & 5.925 & \textbf{5.896} \\
3.07 & 5.980 & \textbf{5.948} \\
2.99 & 6.027 & \textbf{5.989} \\
\bottomrule
\end{NiceTabular}
\end{minipage}\hfill
\begin{minipage}[t]{0.30\linewidth}
\centering
\caption{\KronQ{} scalar vs. HIGGS vector quantizer (L2-7B, ctx 4096).}
\label{tab:higgs}
\vspace{2pt}\footnotesize
\begin{NiceTabular}{lcc}[code-before = \rectanglecolor{gray!15}{2-3}{5-3}]
\toprule
Bits & HIGGS & \KronQ{} \\
\midrule
W4 & 5.213 & \textbf{5.200} \\
W3 & 5.559 & \textbf{5.448} \\
W2 & 8.637 & \textbf{7.603} \\
\bottomrule
\end{NiceTabular}
\end{minipage}
\vspace{-3mm}
\end{table}

\vspace{-1mm}
\section{Why GPTQ and GPTAQ Fail on LLaMA-3-70B}
\label{app:llama3_70b_dist}
\vspace{-7mm}
\begin{figure}[H]
    \centering
    \begin{subfigure}{\linewidth}
        \includegraphics[width=0.95\linewidth]{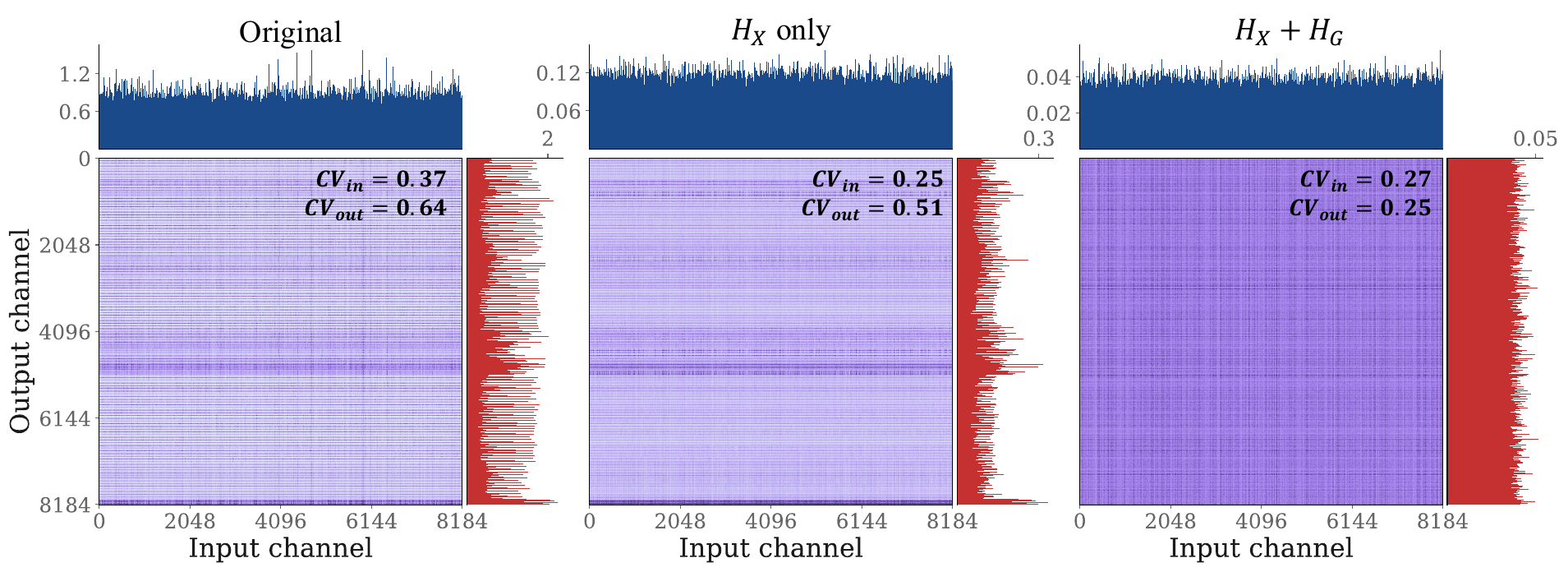}
        \caption{Weight distributions of LLaMA-2-70B}
    \end{subfigure}
    \begin{subfigure}{\linewidth}
        \includegraphics[width=0.95\linewidth]{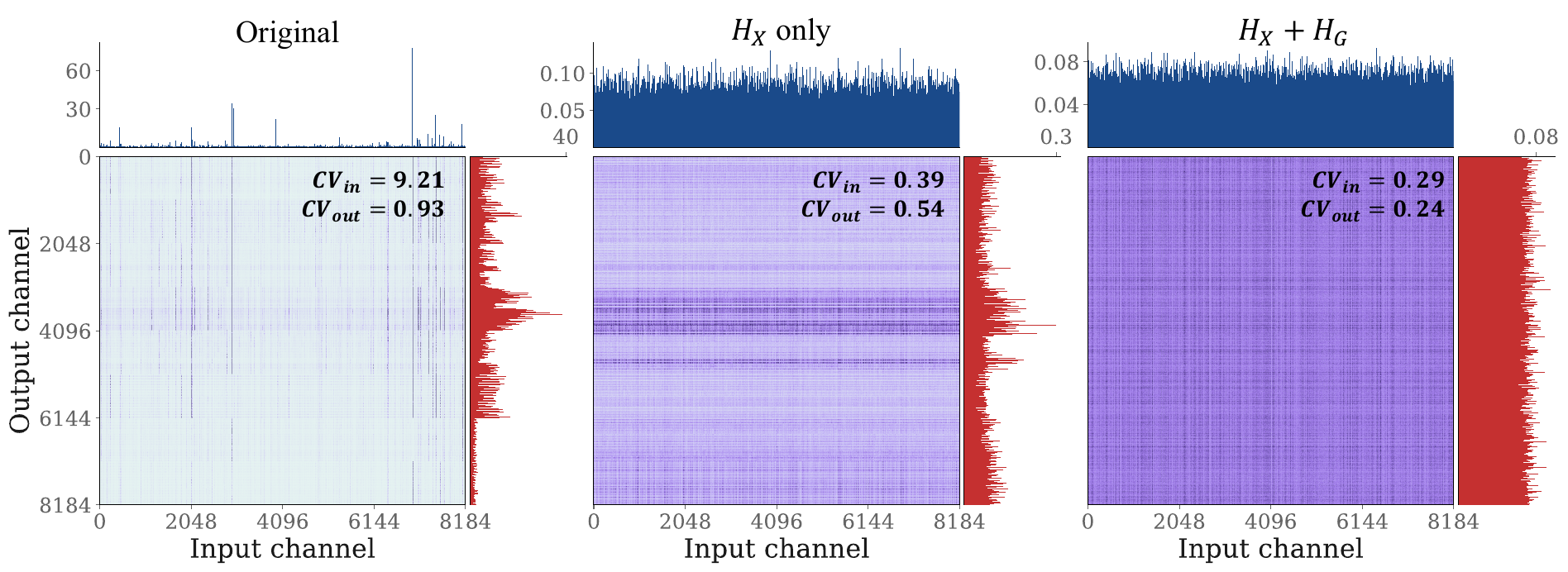}
        \caption{Weight distributions of LLaMA-3-70B}
        \vspace{-1.5mm}
    \end{subfigure}
    \caption{Weight magnitude distribution of the Q projection in the first layer under three configurations: original weights, after column-side incoherence ($\mathbf{H}_X$ only), and after bidirectional incoherence ($\mathbf{H}_X + \mathbf{H}_G$). Top histograms show column norms; right histograms show row norms. $\text{CV}_\text{in}$ and $\text{CV}_\text{out}$ denote the coefficient of variation of column and row norms, respectively.}
    \label{fig:weight_dist}
\end{figure}

Figure~\ref{fig:weight_dist} compares the weight magnitude distributions of LLaMA-3-70B and LLaMA-2-70B. In LLaMA-3-70B, the original weights exhibit extreme column-wise outliers ($\text{CV}_\text{in} = 9.21$) concentrated on specific input dimensions, a phenomenon previously identified by \citet{qin2024uniqueness} as unique to the LLaMA-3/3.1-70B family. These outliers expand the quantization range by orders of magnitude, causing the column-wise OBS updates in GPTQ and GPTAQ to produce degenerate solutions. In contrast, LLaMA-2-70B shows a substantially milder distribution ($\text{CV}_\text{in} = 0.37$), consistent with its robustness to per-channel quantization. Applying $\mathbf{H}_X$-only incoherence suppresses the input-side outliers in LLaMA-3-70B ($\text{CV}_\text{in}: 9.21 \to 0.39$) but leaves the output-side heterogeneity largely unaddressed ($\text{CV}_\text{out}: 0.93 \to 0.54$). BiIP further resolves the output-side structure via $\mathbf{H}_G$, achieving $\text{CV}_\text{in} = 0.29$ and $\text{CV}_\text{out} = 0.24$, comparable to the post-BiIP values of LLaMA-2-70B ($0.27$ and $0.25$).

\section{Per-Sublayer Calibration Memory}
\label{app:memory}
Table~\ref{tab:memory_breakdown} details the memory required to perform calibration
per sublayer for \mbox{LLaMA-2-7B}, LLaMA-3-8B, and LLaMA-2-13B, following the same
convention as \citet{li2025gptaq} (Table~9), with block slices of size $B=128$
included in all reported values.
The Cholesky factor $\mathbf{L}$ is stored as a lower-triangular matrix
and the correction matrix $\mathbf{P}$ as an upper-triangular matrix.
\KronQ{} additionally stores $\mathbf{H}_G$ as a full $d_\mathrm{out} \times d_\mathrm{out}$
matrix during BiIP; once released, the remaining memory footprint is identical to GPTAQ.
The peak memory overhead is therefore transient and confined to the BiIP preprocessing step.

\begin{table}[h]
  \centering
  \caption{Memory needed to perform calibration (GiB) per sublayer.
  $^\dagger$$\mathbf{H}_G$ released after BiIP, remaining memory matches GPTAQ.}
  \label{tab:memory_breakdown}
  \resizebox{\textwidth}{!}{
  \begin{tabular}{llccccccc}
    \toprule
    Model & Method & q\_proj & k\_proj & v\_proj & o\_proj & up\_proj & gate\_proj & down\_proj \\
    \midrule
    \multirow{3}{*}{LLaMA-2-7B}
      & GPTQ   & 0.13 & 0.13 & 0.13 & 0.13 & 0.29 & 0.29 & 0.48 \\
      & GPTAQ  & 0.16 & 0.16 & 0.16 & 0.16 & 0.32 & 0.32 & 0.70 \\
      & \KronQ{}$^\dagger$ & $0.22{\to}0.16$ & $0.22{\to}0.16$ & $0.22{\to}0.16$ & $0.22{\to}0.16$ & $0.77{\to}0.32$ & $0.77{\to}0.32$ & $0.77{\to}0.70$ \\
    \midrule
    \multirow{3}{*}{LLaMA-3-8B}
      & GPTQ   & 0.13 & 0.13 & 0.13 & 0.13 & 0.37 & 0.37 & 0.71 \\
      & GPTAQ  & 0.16 & 0.16 & 0.16 & 0.16 & 0.40 & 0.40 & 1.10 \\
      & \KronQ{}$^\dagger$ & $0.22{\to}0.16$ & $0.22{\to}0.16$ & $0.22{\to}0.16$ & $0.22{\to}0.16$ & $1.17{\to}0.40$ & $1.17{\to}0.40$ & $1.16{\to}1.10$ \\
    \midrule
    \multirow{3}{*}{LLaMA-2-13B}
      & GPTQ   & 0.20 & 0.20 & 0.20 & 0.20 & 0.45 & 0.45 & 0.76 \\
      & GPTAQ  & 0.25 & 0.25 & 0.25 & 0.25 & 0.50 & 0.50 & 1.11 \\
      & \KronQ{}$^\dagger$ & $0.35{\to}0.25$ & $0.35{\to}0.25$ & $0.35{\to}0.25$ & $0.35{\to}0.25$ & $1.22{\to}0.50$ & $1.22{\to}0.50$ & $1.21{\to}1.11$ \\
    \bottomrule
  \end{tabular}}
\end{table}

\section{Limitations}
\label{limit}
\KronQ{} requires a backward pass over the calibration set to estimate the gradient covariance $\mathbf{H}_G$ prior to quantization, introducing additional offline computation relative to activation-only methods such as GPTQ and GPTAQ. Furthermore, $\mathbf{H}_G$ is a full $d_\mathrm{out} \times d_\mathrm{out}$ matrix, incurring additional peak memory during BiIP preprocessing. Nevertheless, once $\mathbf{H}_G$ is precomputed and stored, it incurs no additional computation during quantization itself, as $\mathbf{H}_G$ cancels algebraically in the column-wise updates (Proposition~\ref{prop:hg_cancel}) and is released upon completion of BiIP. During inference, \KronQ{} requires online reversion of the orthogonal transforms $\mathbf{U}$ and $\mathbf{V}$, introducing $\Theta(d_{\mathrm{in}} \log d_{\mathrm{in}} + d_{\mathrm{out}} \log d_{\mathrm{out}})$ overhead per layer. This is identical to QuIP\#~\citep{tseng2024quip} and our fused CUDA kernel keeps the per-layer cost close to that of the $\Theta(d_{\mathrm{in}} d_{\mathrm{out}})$ weight matrix-vector multiply alone.

\end{document}